\documentclass[conference]{IEEEtran}
\usepackage{times}
\usepackage[dvipsnames,table]{xcolor}
\usepackage{xspace}
\usepackage{xparse}
\usepackage{graphicx}
\usepackage[font=small,hypcap=false]{caption}
\usepackage{subcaption}
\usepackage{amsmath, amssymb}
\usepackage{diagbox}
\usepackage{makecell}
\usepackage{booktabs}
\usepackage[numbers,sort&compress]{natbib}
\usepackage{multicol}
\usepackage[
    bookmarks=true,
    colorlinks=true,
    backref=page,
    linkcolor=brown,
    citecolor=RoyalPurple!75,
    urlcolor=Maroon
]{hyperref}
\usepackage{cuted}  
\usepackage{algorithmic}
\usepackage{algorithm}
\usepackage{array} 
\usepackage{tabularx}
\usepackage{tikz}
\usetikzlibrary{calc}

\NewDocumentCommand{\ourshort}{o}{
  \textcolor{Maroon}{\textbf{\texttt{OAT}}}%
  \IfValueT{#1}{\ensuremath{_{#1}}}%
  \xspace
}
\newcommand{\ourlong}{Ordered Action Tokenization\xspace}
\newcommand{\ourlongcolored}{\textcolor{Maroon}{Ordered Action Tokenization\xspace}}
\newcommand{\ourweb}{\href{https://ordered-action-tokenization.github.io/}{ordered-action-tokenization.github.io}}
\newcommand{\fast}{\texttt{FAST}\xspace}
\newcommand{\bin}{\texttt{Bin}\xspace}
\newcommand{\quest}{\texttt{QueST}\xspace}
\newcommand{\diffp}{\texttt{DP}\xspace}
\newcommand{\propertyI}{\textcolor{Plum}{P.1}\xspace}
\newcommand{\propertyII}{\textcolor{Plum}{P.2}\xspace}
\newcommand{\propertyIII}{\textcolor{Plum}{P.3}\xspace}
\newcommand{\scriptstderr}[1]{\textcolor{gray}{{\scriptsize ± #1}}}
\newcommand{\tinystderr}[1]{\textcolor{gray}{{\tiny ± #1}}}

\newcommand{\highlightcell}{\cellcolor{MidnightBlue!10}}

\begin{document}

\title{\textbf{\ourshort}: \ourlong}


\author{
\authorblockN{
Chaoqi Liu$^1$\quad
Xiaoshen Han$^1$\quad
Jiawei Gao$^1$\quad
Yue Zhao$^2$\quad
Haonan Chen$^1$\quad
Yilun Du$^1$
}
\vspace{5pt}
\authorblockA{
$^{1}$Harvard University\quad
$^{2}$Stanford University
}
\vspace{5pt}
\authorblockA{
\ourweb
}
}

\maketitle

\begin{strip}       
    \centering
    \vspace{-40pt}
    \begin{minipage}{\linewidth}
        \centering
        \includegraphics[width=\linewidth]{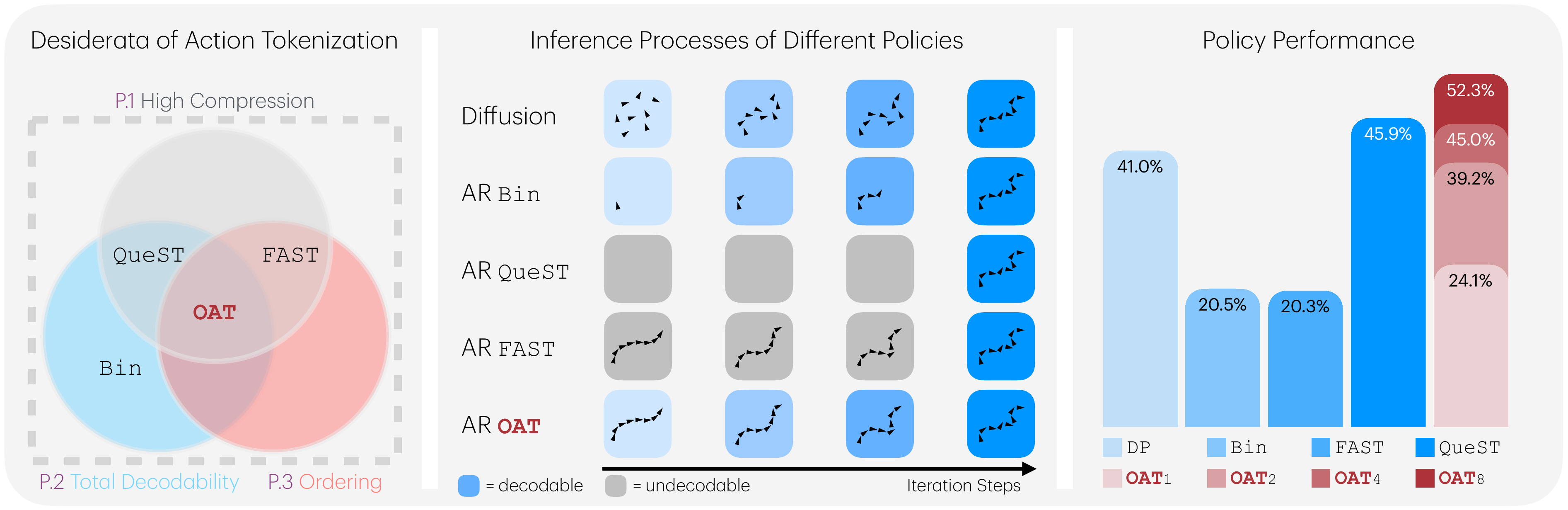}
    \end{minipage}
    \captionof{figure}{\textbf{Left:} Comparison of action tokenization schemes with respect to three desiderata: high compression (\propertyI), total decodability (\propertyII), and left-to-right causally ordered token structure (\propertyIII). Existing methods satisfy only subsets of these properties, while \ourshort uniquely satisfies all three. \textbf{Middle:} Behavior of different policy classes as inference progresses. Diffusion and flow policies refine actions through iterative sampling, while autoregressive policies generate discrete tokens step-by-step. Due to its ordered token space, \ourshort enables prefix-based detokenization: early tokens produce coarse action chunks, and additional autoregressive steps progressively refine actions, enabling flexible, anytime action generation. \textbf{Right:} Overall policy performance aggregated over 20+ tasks.}
    \label{fig:teaser}
    \vspace{-10pt}
\end{strip}

\begin{abstract}
Autoregressive policies offer a compelling foundation for scalable robot learning by enabling discrete abstraction, token-level reasoning, and flexible inference. However, applying autoregressive modeling to continuous robot actions requires an effective action tokenization scheme. Existing approaches either rely on analytical discretization methods that produce prohibitively long token sequences or learned latent tokenizers that lack structure, limiting their compatibility with next-token prediction. In this work, we identify three desiderata for action tokenization — high compression, total decodability, and a left-to-right causally ordered token space — and introduce \textit{\ourlongcolored} (\ourshort), a learned action tokenizer that satisfies all three. \ourshort discretizes action chunks into an ordered sequence of tokens using a transformer with registers, finite scalar quantization, and ordering-inducing training mechanisms. The resulting token space aligns naturally with autoregressive generation and enables prefix-based detokenization, yielding an anytime trade-off between inference cost and action fidelity. Across more than 20 tasks spanning four simulation benchmarks and real-world settings, autoregressive policies equipped with \ourshort consistently outperform prior tokenization schemes and diffusion-based baselines, while offering significantly greater flexibility at inference time.
\end{abstract}

\IEEEpeerreviewmaketitle

\section{Introduction}

Autoregressive sequence models have emerged as a powerful foundation for modern robot learning. In particular, large transformer-based policies have demonstrated strong generalization when trained directly on robotic data~\cite{rt12022arxiv, octo2023} or adapted from pre-trained vision-language backbones~\cite{kim24openvla, rt22023arxiv, wen2025tinyvla}. A critical but often under-examined component underlying these successes is how continuous robot actions are represented as discrete symbols suitable for autoregressive generation.

This representation problem is known as \textit{action tokenization}: the process of mapping continuous control signals into a sequence of discrete tokens. Experience from natural language processing and computer vision has shown that tokenization is far more than an implementation detail — it fundamentally shapes learning dynamics, model capacity utilization, scalability, and downstream performance~\cite{sennrich2016neural, xue2022byt5tokenfreefuturepretrained, dosovitskiy2021imageworth16x16words, bao2022beitbertpretrainingimage}. Despite its centrality, action tokenization for robot control remains significantly less understood than its counterparts in language and vision.

The dominant approach in existing autoregressive robot policies relies on naive discretization via per-dimension binning~\cite{rt12022arxiv, rt22023arxiv, kim24openvla}. While conceptually simple, this strategy yields extremely long token sequences whose lengths scale linearly with action dimensionality and prediction horizon, leading to substantial inefficiencies in training and inference. To alleviate this issue, recent work has explored learned latent tokenizers~\cite{mete2024quest, belkhale2024minivla, lee2024behaviorgenwithlatentactions} and analytical compression methods such as frequency-domain compression~\cite{pertsch2025fast}. However, these alternatives introduce their own limitations: learned tokenizers often produce unstructured latent spaces that are poorly aligned with next-token prediction, while existing frequency-domain approaches may sacrifice decodability. Across these approaches, we identify a persistent and fundamental limitation: existing action tokenization strategies face an inherent trade-off between compression rate, modelability\footnote{\textit{Modelability} characterizes how challenging it is for generative models to capture the distribution of the representation~\cite{kolesnikov2022uvim, dieleman2025latents,kim2025trainworstplanbest}.} under autoregressive learning, and decodability. Improving one aspect typically degrades another, resulting in token spaces that are either too long to model efficiently, insufficiently structured for stable generation, or partially decodable at inference time.

In this work, we argue that an effective action tokenizer for autoregressive policies must simultaneously satisfy three key properties (Fig.~\ref{fig:teaser} left): \textbf{(\propertyI) High Compression}, reducing the effective prediction horizon to enable efficient long-context modeling; \textbf{(\propertyII) Total Decodability}, meaning the decoder is a total function in which every token sequence maps to a valid action chunk, with no undefined or invalid outputs; and \textbf{(\propertyIII) Causal Ordering}, imposing a left-to-right structure over tokens that aligns with the inductive bias of next-token prediction. While prior methods satisfy subsets of these desiderata, none achieve all three simultaneously.

To bridge this gap, we introduce \textit{\ourlongcolored} (\ourshort), a learned action tokenizer that discretizes continuous action chunks into highly compressed and \textit{causally ordered} token sequences. \ourshort employs transformer-based register tokens to aggregate temporal information, finite scalar quantization (FSQ) to construct a discrete bottleneck, and nested dropout to explicitly induce ordering that aligns the latent space with autoregressive generation. The resulting tokenization ensures that any token prefix corresponds to a plausible action chunk. Beyond improved modelability, the ordered structure learned by \ourshort enables a key capability absent from prior approaches: \textit{prefix-based decoding}. Autoregressive policies may terminate generation early and still produce valid actions, yielding a natural trade-off between computation and action fidelity. As additional tokens are generated, decoded actions are progressively refined.

In summary, this paper makes three contributions: \textbf{(i)} we formalize a set of necessary desiderata for action tokenization in autoregressive robot policies, exposing a fundamental trade-off faced by existing methods; \textbf{(ii)} we propose \ourshort, a novel tokenizer that uniquely satisfies compression, total decodability, and causal ordering simultaneously; and \textbf{(iii)} we demonstrate that \textit{ordering} is the critical ingredient for stable and scalable autoregressive learning, enabling superior performance and flexible, prefix-based decoding across 20+ simulation and real-world manipulation tasks.

\section{Related Work on Generative Policies}

We focus on policies of the form $\pi(a_{1:H_a} \mid o_{1:H_o})$ that predict a \textit{chunk} of actions conditioned on a history of observations. Predicting multi-step action sequences has been shown to improve temporal consistency, reduce compounding error, and stabilize long-horizon behavior compared to single-step prediction~\cite{zhao2023learningfinegrainedbimanualmanipulation, chi2024diffusionpolicy, zhang2025actionchunkingexploratorydata}. Action chunking also amortizes inference cost over multiple time steps and has become a standard design choice in modern robot policies.

Diffusion and flow-based policies~\cite{reuss2023goalconditionedimitlearning, hou2025diffusiontransformerpolicy, wolf2025diffusionmodelsroboticmanipulation, høeg2025hybriddiffusionsimultaneoussymbolic, tie2025etseed, yang2025diffusionmodelscomprehensivesurvey, chen2025multimodalmanipmultimodalpolicy, chen2025bimanual, zhang2024trajflow, intelligence2025pi05vla, liu2025factorpolicy, chen2025tool, chi2024diffusionpolicy, janner2022diffuser, liu2025manipulation, xiong2025via} have proven highly effective for continuous action generation and imitation learning, and are widely used as standalone robot policies. More recently, in VLA systems, diffusion and flow models are increasingly employed as \textit{action experts} or continuous decoding heads that translate higher-level representations into executable actions~\cite{intelligence2025pi05vla, wen2025tinyvla, liu2025hybridvla, nvidia2025gr00tn1, black2024pi0vla}. In this role, they complement discrete reasoning and planning components by providing expressive, high-fidelity action synthesis.

Autoregressive policies model the distribution of action sequences by factorizing it into a product of conditional distributions, generating one element at a time~\cite{vaswani2017attnisallyouneed}. Autoregressive models have demonstrated remarkable scalability and generalization in language, image, and video modeling~\cite{radford2019langmodelareunsupervisedmultitasklearners, touvron2023llamaopenefficientfoundation, yang2025qwen3technicalreport, wang2025hmavideo, li2024mar}. This success has motivated their adoption in robotics, particularly within VLA systems~\cite{rt12022arxiv, rt22023arxiv, octo2023, kim24openvla, nvidia2025gr00tn1, wen2025tinyvla, pertsch2025fast, oneill2024oxe, huang2025roboground}. 

Despite their success, the effectiveness of autoregressive policies in robotics depends critically on the choice of tokenization. In this work, we systematically study the key desiderata of action tokenization for autoregressive policies and propose a principled tokenizer that addresses these requirements. We formalize these properties and introduce \ourshort in the following sections.

\begin{figure*}[t]
    \centering
    \begin{subfigure}[t]{0.19\linewidth}
        \centering
        \caption{\ourshort[1] $\mathrm{MSE}=0.592$\vspace{2.5pt}}
        \includegraphics[width=\linewidth]{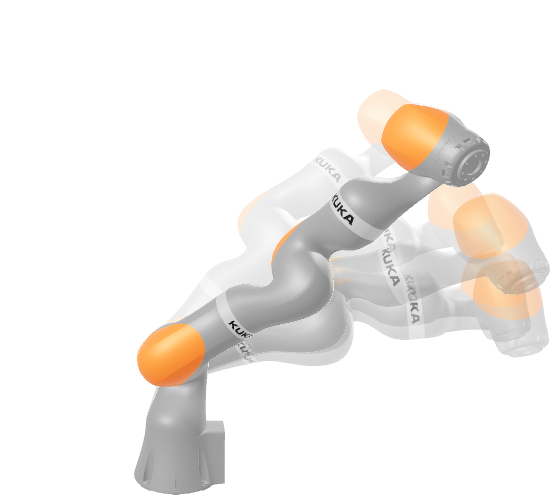}
    \end{subfigure}
    \hfill
    \begin{subfigure}[t]{0.19\linewidth}
        \centering
        \caption{\ourshort[2] $\mathrm{MSE}=0.446$\vspace{2.5pt}}
        \includegraphics[width=\linewidth]{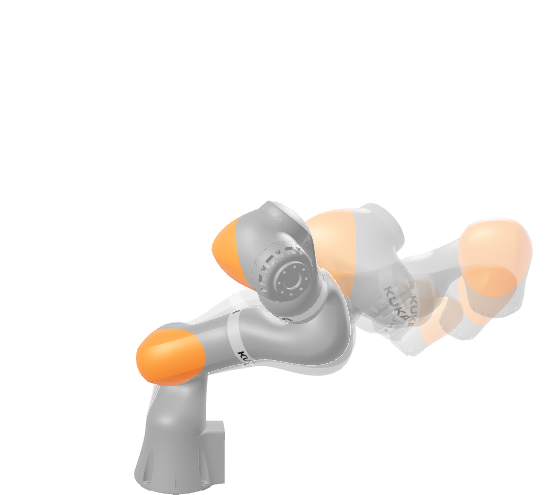}
    \end{subfigure}
    \hfill
    \begin{subfigure}[t]{0.19\linewidth}
        \centering
        \caption{\ourshort[4] $\mathrm{MSE}=0.038$\vspace{2.5pt}}
        \includegraphics[width=\linewidth]{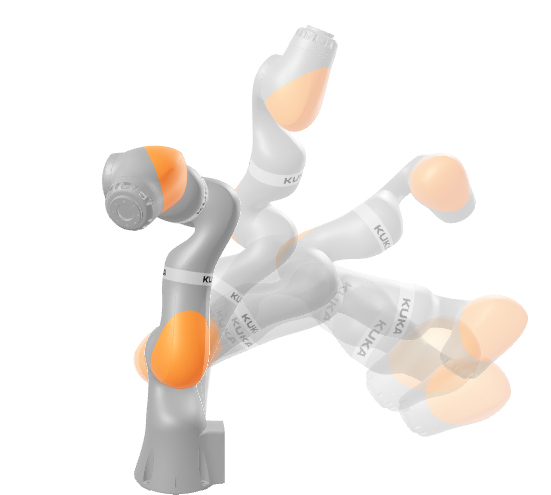}
    \end{subfigure}
    \hfill
    \begin{subfigure}[t]{0.19\linewidth}
        \centering
        \caption{\ourshort[8] $\mathrm{MSE}=0.009$\vspace{2.5pt}}
        \includegraphics[width=\linewidth]{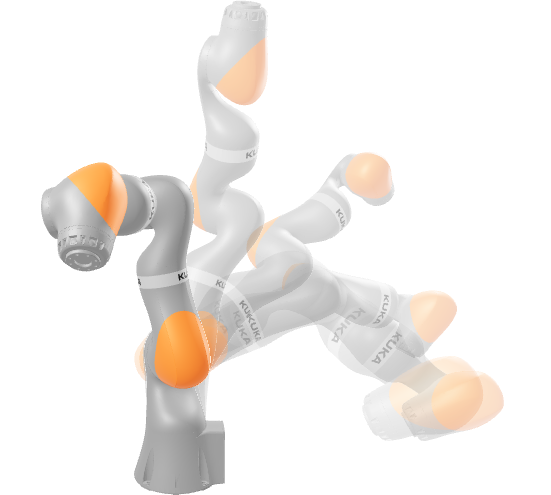}
    \end{subfigure}
    \hfill
    \begin{subfigure}[t]{0.19\linewidth}
        \centering
        \caption{Ground Truth\vspace{2.5pt}}
        \includegraphics[width=\linewidth]{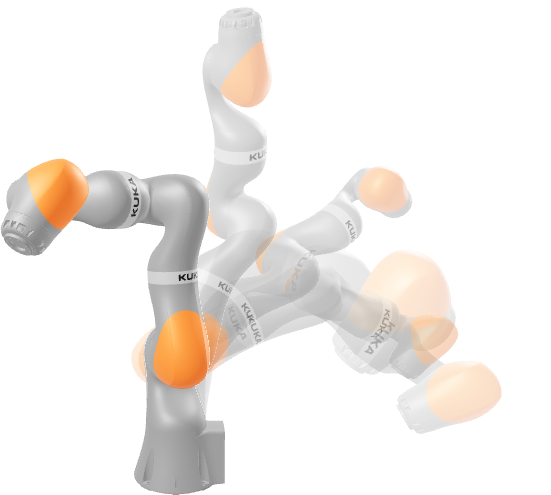}
    \end{subfigure}
    \caption{\textbf{Coarse-to-fine action chunk reconstruction.} Visualization of reconstructed action chunks using increasing numbers of decoded tokens. Panels (a–d) show \ourshort reconstructions using $K \in \{1,2,4,8\}$ tokens, respectively, while (e) shows the ground-truth action chunk. Earlier tokens capture the coarse, global structure of the motion, while additional tokens progressively refine fine-grained details, yielding trajectories that increasingly match the ground truth. Ghosted poses indicate temporal progression within each reconstructed action chunk. Interactive visualization on project website: \ourweb.}
    \label{fig:our_tok_iiwa_rollout_ghosted}
    \vspace{-15pt}
\end{figure*}

\section{Action Tokenization Preliminaries}
\label{sec:action_token_prelim}

Robot actions, however, are inherently continuous and high-dimensional. To enable autoregressive modeling, continuous action chunks must first be discretized into a sequence of tokens. This process, referred to as \textit{action tokenization}, defines a mapping
\begin{equation*}
\mathcal{T}: a_{1:H_a} \;\mapsto\; T_{1:H_l},
\end{equation*}
which maps a continuous action chunk of horizon $H_a$ and dimensionality $D_a$ to a sequence of $H_l$ discrete tokens drawn from a vocabulary $\mathcal{V}$. A corresponding \textit{detokenization} mapping
\begin{equation*}
\mathcal{T}^{-1}: T_{1:H_l} \;\mapsto\; a_{1:H_a}
\end{equation*}
maps token sequences back into continuous action space, producing executable action chunks. Autoregressive policies operate entirely in the discrete token space defined by $\mathcal{T}$, while control execution relies on $\mathcal{T}^{-1}$ to convert generated token sequences into continuous actions.

We argue that an efficient and effective action tokenizer, that balances \textit{rate-distortion-modelability trade-off}~\cite{shannon1948maththeoryofcommunication, tschannen2018recentadvvaerepr, blau2019rethinkinglossycomprratedistortionpercep, dieleman2025latents, zhao2025npq}, should satisfy the following three properties:
\begin{itemize}
    \item[] \textbf{\propertyI $\mathcal{T}$ achieves a high compression rate.}
    \item[] \textbf{\propertyII $\mathcal{T}^{-1}$ is a well-defined total function.}
    \item[] \textbf{\propertyIII $T_{1:H_l}$ has a left-to-right causal ordering.}
\end{itemize}
\noindent \propertyI: The token horizon $H_l$ should be sufficiently small to enable efficient autoregressive modeling, while retaining enough capacity to preserve necessary information from the original action chunk. \propertyII: The decoder $\mathcal{T}^{-1}$ must be a well-defined \textit{total function}: for every token sequence $T_{1:H_l}$ in the discrete token space, $\mathcal{T}^{-1}(T_{1:H_l})$ produces a valid action chunk $a_{1:H_a}$. This property is critical in autoregressive settings, where policies may generate arbitrary token sequences at inference time. If $\mathcal{T}^{-1}$ is only partially defined, invalid or non-decodable token sequences can lead to undefined behavior and catastrophic failures during execution. \propertyIII: The token sequence $T_{1:H_l}$ should admit a meaningful left-to-right causal ordering aligned with causal, next-token prediction. Such a structure is essential for stable autoregressive generation: early tokens should capture coarse, globally salient aspects of the action chunk, while later tokens refine finer details. An ordered token space improves controllability, robustness, and compatibility with prefix-based generation, and we revisit this property throughout the paper both conceptually and empirically.

\subsection{Binning}
The most commonly used action tokenization approach is per-dimension binning (\bin)~\cite{rt12022arxiv, rt22023arxiv, kim24openvla}. For each action dimension, the range of values observed in the dataset is normalized to $[-1,1]$, then divided into $N$ uniform bins, and each continuous value is mapped to its corresponding bin index. Given an action chunk of shape $H_a \times D_a$, binning produces a token sequence
\begin{equation*}
    \mathcal{T}(a_{1:H_a}) = [T_{1,1}, ..., T_{1,D_a}, T_{2,1}, ..., T_{H_a,D_a}],
    \quad T_{i,j} \in [N].
\end{equation*}

While \bin is conceptually simple and yields a well-defined, totally decodable mapping (\propertyII), it does not provide the left-to-right ordering we seek (\propertyIII): the token order is a serialization over dimensions and time rather than a hierarchy aligned with causal next-token prediction. Moreover, \bin scales poorly — long horizons and high-dimensional actions can produce hundreds of tokens per chunk — severely slowing training and inference and introducing substantial latency. Therefore, \bin fails to satisfy \propertyI and \propertyIII, despite meeting \propertyII.

\subsection{Frequency-domain Transform}
An alternative line of work explores frequency-domain compression, for instance \textit{Frequency-space Action Sequence Tokenization} (\fast)~\cite{pertsch2025fast}, which employs the Discrete Cosine Transform (DCT) to decompose action chunks into frequency coefficients, followed by Byte Pair Encoding (BPE)~\cite{gage1994bpe}. \fast achieves high information density (\propertyI), and crucially, its low-frequency components first then high-frequency components ordering (\propertyIII) improves downstream autoregressive policies: early token predictions capture the overall trajectory shape, stabilizing rollout before finer details are generated.

However, \fast detokenization $\mathcal{T}^{-1}$ is a partial function that violates \propertyII. Because BPE produces variable-length sequences, there is no guarantee that an arbitrary token sequence generated by the policy will decode into a valid frequency matrix of fixed dimensions. This structural mismatch renders the decoding function partially undefined for invalid token counts, leading to potential runtime failures. We refer readers to Appendix~\ref{appendix:sec:fast_decoding_error} and the discussion on Hugging Face\footnote{\href{https://huggingface.co/physical-intelligence/fast/discussions/4}{https://huggingface.co/physical-intelligence/fast/discussions/4}} for further details.

\subsection{Quantized Latents}
Another line of work explores learned compression via encoder-decoder architectures with vector quantization~\cite{mete2024quest, lee2024behaviorgenwithlatentactions, belkhale2024minivla}. These methods map action chunks into a latent sequence of shape $H_l \times D_l$, which is quantized~\cite{oord2017vq, mentzer2024fsq} into tokens. The latent horizon $H_l$ and dimension $D_l$ are hyperparameters, often chosen relative to $H_a$ and $D_a$. Such approaches can achieve extremely high compression; for example, mapping action chunks of horizon $H_a = 32$ into latent sequences with $H_l = 8$ tokens, satisfying \propertyI. Because $\mathcal{T}$ and $\mathcal{T}^{-1}$ are approximated by encoder and decoder neural networks respectively, $\mathcal{T}^{-1}$ are always total (\propertyII).

However, existing learned tokenizers typically produce unstructured token spaces. The tokens lack a consistent ordering or hierarchical abstraction, making them poorly suited for autoregressive generation. As a result, while existing learned tokenizers satisfy \propertyI and \propertyII, they fail to meet \propertyIII.

\section{\texorpdfstring{\ourshort: \ourlongcolored}{\ourlongcolored}}

\begin{figure*}[t]
    \centering
    \includegraphics[width=\linewidth]{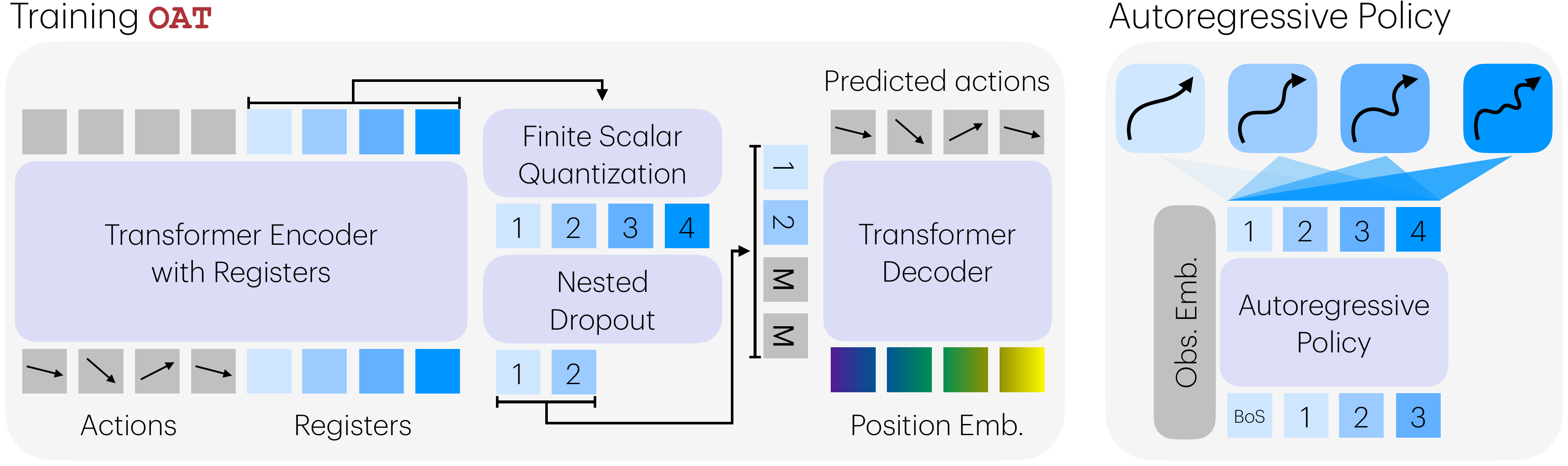}
    \caption{\textbf{\ourshort overview.} \textbf{Left:} \ourshort maps a chunk of continuous actions into an ordered sequence of discrete tokens using a transformer encoder with register tokens, FSQ, and nested dropout to induce token ordering. The resulting tokens form a compact action representation, which is decoded to reconstruct action chunks for downstream autoregressive policies. \textbf{Right:} During \ourshort policy inference, tokens are generated autoregressively and can be detokenized from any prefix. As more autoregressive steps are taken, additional tokens progressively refine the decoded action chunk, producing actions with increasing temporal and spatial detail. \ourshort enables flexible, anytime action generation.}
    \label{fig:our_method_overview}
    \vspace{-15pt}
\end{figure*}

Our objective is to learn an action tokenizer that satisfies three desiderata introduced in Sec.~\ref{sec:action_token_prelim}: high compression (\propertyI), total decodability (\propertyII), and a structured ordering over tokens (\propertyIII). While prior learned tokenizers achieve compact and decodable representations, they lack an explicit ordering over latent tokens~\cite{mete2024quest, lee2024behaviorgenwithlatentactions, belkhale2024minivla}, which limits their compatibility with autoregressive policies. We introduce \ourshort, a learned autoencoder framework that discretizes action chunks into an ordered sequence of tokens. \ourshort encodes actions using transformer-based register tokens, discretizes the resulting latents with FSQ~\cite{mentzer2024fsq}, and reconstructs actions via a conditional decoder. To induce ordering in the token space, we combine causal attention over register tokens with nested dropout during training. Together, these design choices encourage an ordered latent representation in which earlier tokens capture coarse, global structure and later tokens refine details (Fig.~\ref{fig:our_tok_iiwa_rollout_ghosted}). As a result, \ourshort supports decoding from any prefix of the token sequence, enabling variable-length and anytime reconstruction of action chunks. We demonstrate \ourshort training pipeline (left) and autoregressive policies operate on \ourshort (right) in Fig.~\ref{fig:our_method_overview}.

\subsection{Tokenization \texorpdfstring{$\mathcal{T}$}{T} and Detokenization \texorpdfstring{$\mathcal{T}^{-1}$}{invT}}

\begin{algorithm}[t]
\caption{\ourshort Tokenizer Training}
\label{algo:train_our_tok}
\begin{algorithmic}[1]
\REQUIRE Dataset $\mathcal{D}$ of action chunks $\{a_{1:H_a}\}$; encoder $E_\phi(\cdot)$; learnable register tokens $r_{1:H_l}$; quantizer $\mathrm{FSQ}(\cdot)$; decoder $D_\theta(\cdot)$; learnable mask token $\mathtt{MASK}$; nested-dropout distribution $p(\cdot)$.
\WHILE{not converged}
    \STATE Sample action chunk $a_{1:H_a} \sim \mathcal{D}$
    \STATE Encoding: $\tilde{a}_{1:H_a} \oplus z_{1:H_l} \gets E_\phi(a_{1:H_a} \oplus r_{1:H_l})$
    \STATE Quantization: $\hat{z}_{1:H_l} \gets \mathrm{FSQ}(z_{1:H_l})$
    \STATE Tail dropout: $\hat{z}_{1:H_l} \gets \hat{z}_{1:K} \oplus \langle\mathtt{MASK}\rangle_{K+1:H_l}, K \sim p(\cdot)$
    \STATE Decoding: $\hat{a}_{1:H_a} \gets D_\theta(\hat{z}_{1:H_l})$
    \STATE Reconstruction loss: $\mathcal{L} \gets \| \hat{a}_{1:H_a} - a_{1:H_a} \|_2^2$
    \STATE Optimization: $\{ \phi, r, \theta, \mathtt{MASK} \} \gets \{ \phi, r, \theta, \mathtt{MASK} \} - \eta \nabla \mathcal{L}$
\ENDWHILE
\STATE $\mathcal{T}(\cdot) \gets \{ E_\phi, r_{1:H_l}, \mathrm{FSQ} \}$,\;
       $\mathcal{T}^{-1}(\cdot) \gets \{ D_\theta, \mathtt{MASK} \}$
\RETURN $\mathcal{T}(\cdot), \mathcal{T}^{-1}(\cdot)$
\end{algorithmic}
\end{algorithm}

The objective of the tokenizer $\mathcal{T}$ is to compress a continuous action chunk of shape $H_a \times D_a$ into a compact discrete representation of shape $H_l \times D_l$. To this end, we concatenate the input action sequence with a fixed set of learnable \textit{register tokens}, $r_{1:H_l}$, which act as a compact read--write memory for summarizing the temporal structure of the input~\cite{darcet2024vitneedsregisters, yu2024imageworth32tok}. A transformer encoder jointly processes the action chunk and register tokens, allowing information from the action sequence to be aggregated into the registers. After encoding, the register tokens form the bottleneck representation~\cite{dieleman2025latents} of the autoencoder, while the encoded action tokens are discarded.

The register latents $z_{1:H_l}$ are discretized using FSQ, yielding a sequence of $H_l$ discrete tokens $T_{1:H_l}$. These tokens constitute the action representation used both for reconstruction during tokenizer training and as the action space for downstream autoregressive policies.

The decoder implements the detokenization mapping $\mathcal{T}^{-1}$, generating a continuous action chunk conditioned on the discrete token sequence. The \ourshort framework imposes no restrictions on the specific decoder architecture or training objective. In this work, we employ a single-pass transformer decoder similar to~\cite{zhao2023actpolicy} (see Fig.~\ref{fig:our_method_overview}), which we find provides a favorable trade-off between reconstruction quality, stability, and computational efficiency. We provide more details on decoding in Appendix~\ref{appendix:sec:our_tok_dec}. The tokenizer $\mathcal{T}$ and detokenizer $\mathcal{T}^{-1}$ are trained jointly end-to-end using a reconstruction objective. Pseudocode for \ourshort training is provided in Algo.~\ref{algo:train_our_tok}, also see Fig.~\ref{fig:our_method_overview} for the pipeline.

\subsection{Inducing Token Ordering For Modelability}
Prior work has highlighted the importance of left-to-right causal ordering for effective autoregressive modeling~\cite{kolesnikov2022uvim, khemakhem2021causalarxflow, im2025deepautoregressivemodelscausal, kim2025trainworstplanbest}. To align the learned token space with autoregressive generation, we explicitly induce a left-to-right ordering over tokens $T_{1:H_l}$ that naturally aligns with next-token prediction similar to \cite{bachmann2025flextok, wen2025principalcomponentsenablenew, ripple2014learnorderedreprwithnesteddropout}. Our goal is to ensure that earlier tokens capture coarse, globally salient aspects of an action chunk, while later tokens refine finer details. We introduce two complementary mechanisms to impose an ordering and support variable-length token sequences.

\subsubsection{Nested Dropout} 
We train \ourshort to produce an ordered representation by applying nested dropout to the register tokens during training~\cite{cai2025matryoshkamultimodal, kusupati2022matryoshkareprlearning, ripple2014learnorderedreprwithnesteddropout, bachmann2025flextok}. Given register tokens of length $H_l$, we randomly sample the number of tokens to retain, $K \in [H_l]$, and mask out the remaining $H_l - K$ tail tokens. Under this training regime, the encoder is encouraged to pack information into the register tokens in a prioritized, ordered manner, while the decoder learns to reconstruct action chunks from variably sized token prefixes. As a result, the first few tokens capture the most important aspects of the action sequence, while additional tokens progressively refine the reconstruction. Simple action chunks can therefore be faithfully represented with few tokens, whereas more complex behaviors require longer token sequences. Importantly, this ordering is not manually specified but emerges naturally from the nested dropout objective applied to the register tokens.

\subsubsection{Causal Attention} 
Complementary to nested dropout, we impose a causal attention~\cite{vaswani2017attnisallyouneed} structure over the register tokens to further reinforce ordering. Specifically, the encoded action tokens attend freely to one another but do not attend to registers. Each register token attends to all action tokens, enabling global aggregation, but register-register attention is causally masked such that the $i$-th register token only attends to the $j$-th register token if $i \geq j$. This causal dependency structure enforces a left-to-right information flow among registers~\cite{bachmann2025flextok}, aligning the learned token sequence with autoregressive prediction and stabilizing generation from partial prefixes.

\subsection{Information-Theoretic Interpretation of Token Ordering}

The ordering induced by \ourshort admits a natural interpretation from information theory. Classical results by Shannon show that the optimal code length for representing an event scales with the negative logarithm of its probability, i.e., $-\log p$~\cite{shannon1948maththeoryofcommunication}: frequent patterns require fewer bits to encode, while rare or atypical events demand greater representational capacity. In our setting, action chunks $a_{1:H_a}$ are drawn from a data distribution with highly non-uniform structure — most trajectories share common coarse patterns, while fine-grained deviations occur less frequently.

Under this lens, the ordered token sequence $T_{1:H_l}$ learned by \ourshort can be viewed as an implicit progressive coding of action information. Early tokens are encouraged to capture the dominant motion pattern shared across many trajectories. Later tokens then progressively correct residual errors and local details. This behavior emerges naturally from nested dropout: since prefixes must reconstruct actions under partial information, the tokenizer learns to allocate information in decreasing order of frequency and importance. This interpretation explains both the monotonic improvement in reconstruction quality with increasing prefix length and the strong alignment between token order and autoregressive next-token prediction. Importantly, the ordering is not imposed heuristically but arises from optimizing reconstruction under variable information budgets.

\subsection{Autoregressive \texorpdfstring{\ourshort}{} Policies}

\begin{algorithm}[t]
\caption{Autoregressive \ourshort Policy Inference}
\label{algo:our_policy_infer}
\begin{algorithmic}[1]
\REQUIRE Observation history $o_{1:H_o}$; autoregressive policy $\pi(\cdot)$; detokenizer $\mathcal{T}^{-1} = \{ D(\cdot), \mathtt{MASK} \}$; prefix length $K \leq H_l$.
\STATE Initialize empty token prefix $T_{1:K} \gets \varnothing$
\FOR{$i \gets 1$ \TO $K$}
    \STATE Next-token sampling: $T_i \sim \pi(\,\cdot \mid T_{<i}, o_{1:H_o}\,)$
    \STATE $T_{1:K} \gets T_{1:K} \oplus T_i$
\ENDFOR
\STATE Pad tail tokens: $T_{1:H_l} \gets T_{1:K} \oplus \langle\mathtt{MASK}\rangle_{K+1:H_l}$
\STATE Detokenize to action chunk: $\hat{a}_{1:H_a} \gets \mathcal{T}^{-1}(T_{1:H_l})$
\RETURN $\hat{a}_{1:H_a}$
\end{algorithmic}
\end{algorithm}

We use \ourshort as the discrete action representation for autoregressive policy learning. Given an observation history $o_{1:H_o}$, the policy models a distribution over action tokens by factorizing \begin{equation*}
p(T_{1:H_l} \mid o_{1:H_o}) = \prod_{i=1}^{H_l} p(T_i \mid T_{<i}, o_{1:H_o}),
\end{equation*} and generates tokens sequentially. The resulting token sequence is detokenized via $\mathcal{T}^{-1}$ to produce a continuous action chunk for execution.

The ordered token space (\propertyIII) induced by \ourshort is essential for effective autoregressive modeling. Earlier tokens encode the coarse, global structure of the action chunk, while later tokens progressively refine finer details, aligning next-token prediction with the semantics of action generation. As a result, prefixes of the token sequence correspond to valid, increasingly detailed action chunks rather than arbitrary partial reconstructions. 

Crucially, autoregressive generation need not proceed to completion. Because any prefix $T_{1:K}$ can be detokenized into a valid action chunk, \ourshort supports prefix-based execution and enables an anytime trade-off between computation and performance. Short prefixes yield fast but coarse predictions, while longer prefixes produce more refined actions at higher computational cost. This flexibility arises naturally from the ordered tokenization and requires no changes to the policy architecture or training objective, distinguishing \ourshort from prior tokenizers that rely on fixed-length detokenization. The pseudocode for \ourshort policy inference is provided in Algo.~\ref{algo:our_policy_infer}.

\begin{figure*}[t]
    \centering
    \begin{subfigure}[t]{0.48\linewidth}
        \centering
        \caption{LIBERO}
        \includegraphics[width=\linewidth]{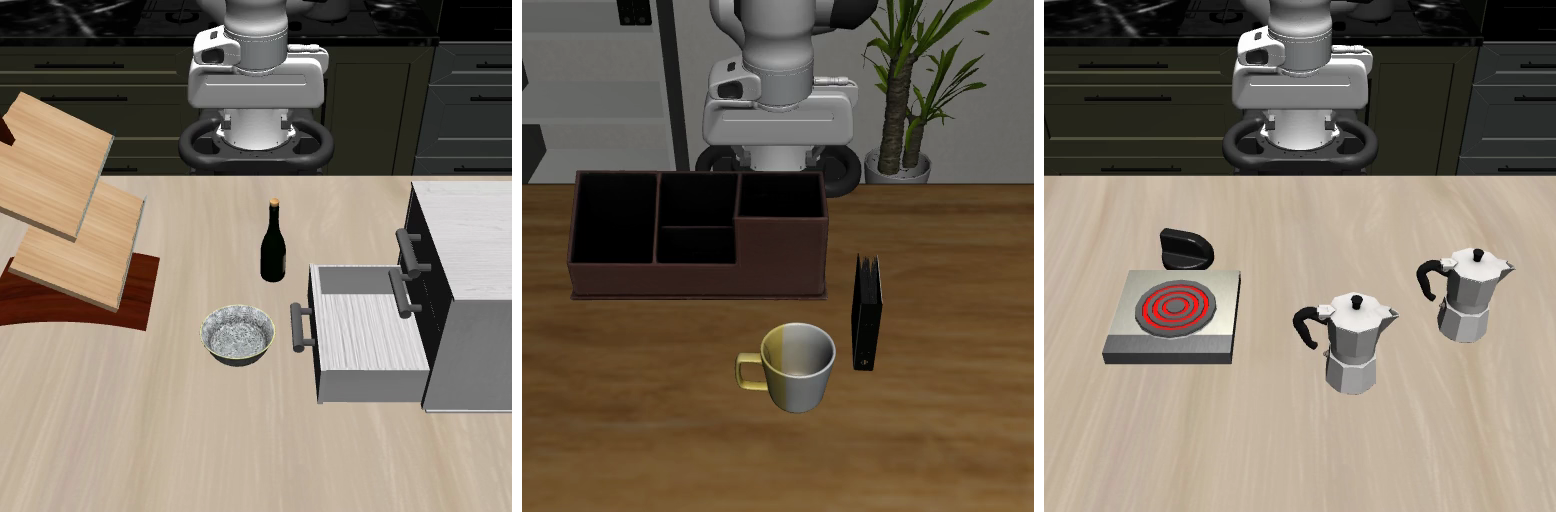}
    \end{subfigure}
    \hspace{2.5pt}
    \begin{subfigure}[t]{0.48\linewidth}
        \centering
        \caption{RoboMimic}
        \includegraphics[width=\linewidth]{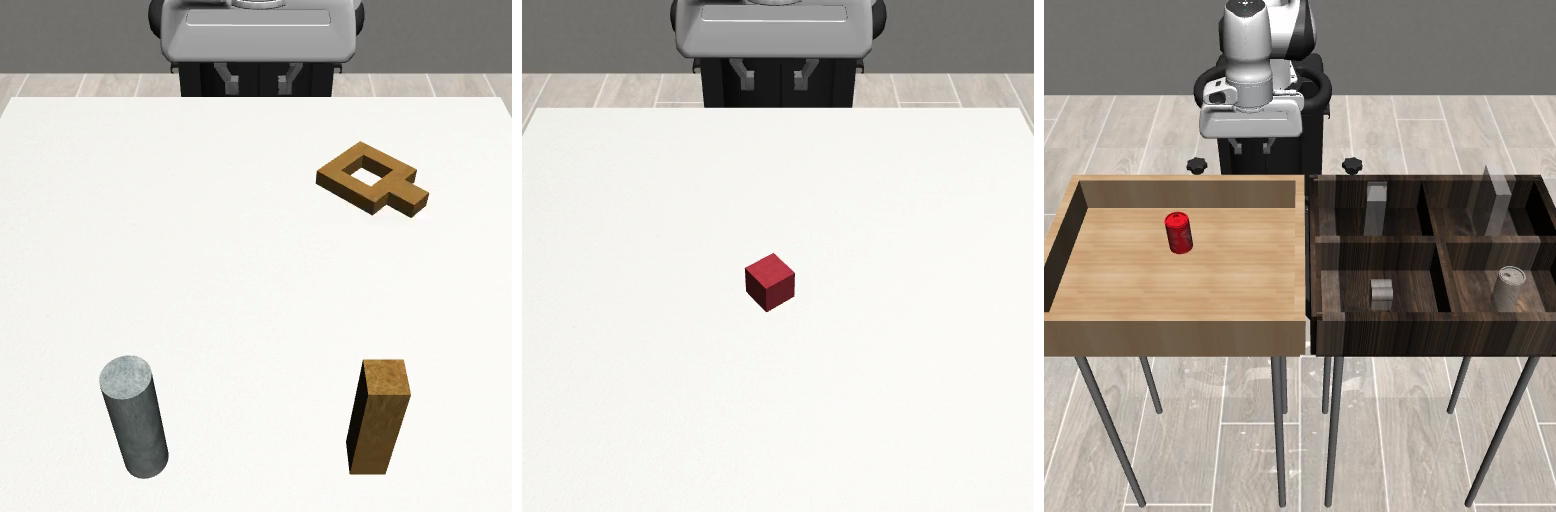}
    \end{subfigure}
    \begin{subfigure}[t]{0.48\linewidth}
        \centering
        \caption{MetaWorld}
        \includegraphics[width=\linewidth]{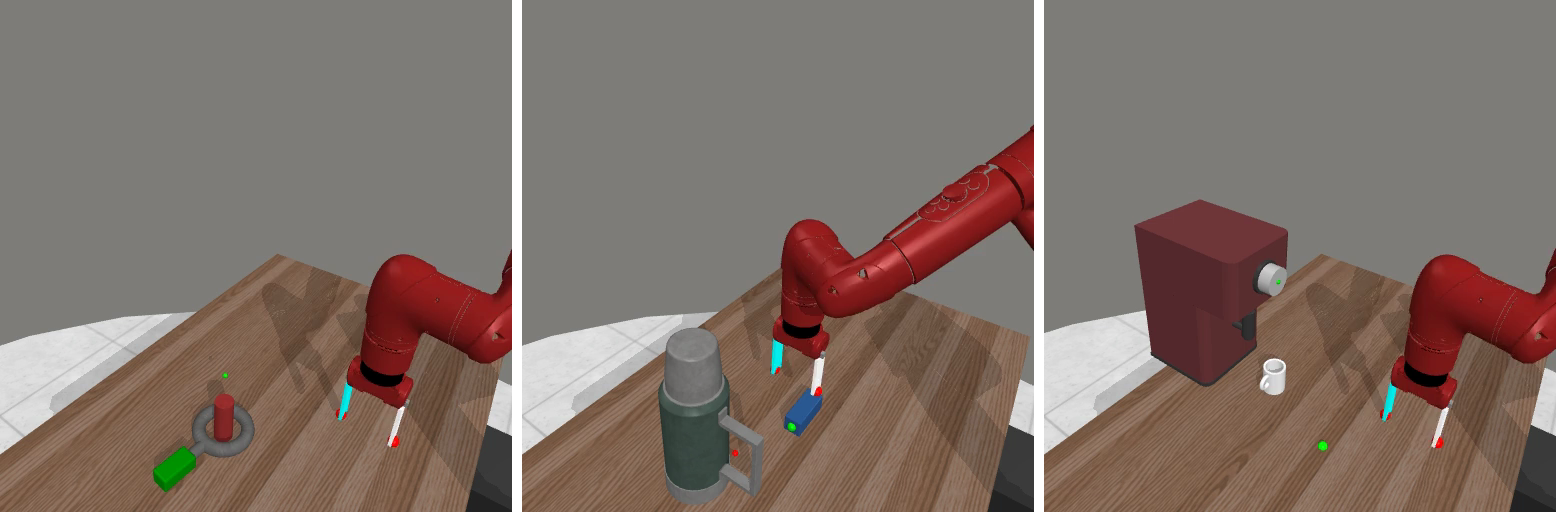}
    \end{subfigure}
    \hspace{2.5pt}
    \begin{subfigure}[t]{0.48\linewidth}
        \centering
        \caption{RoboCasa}
        \includegraphics[width=\linewidth]{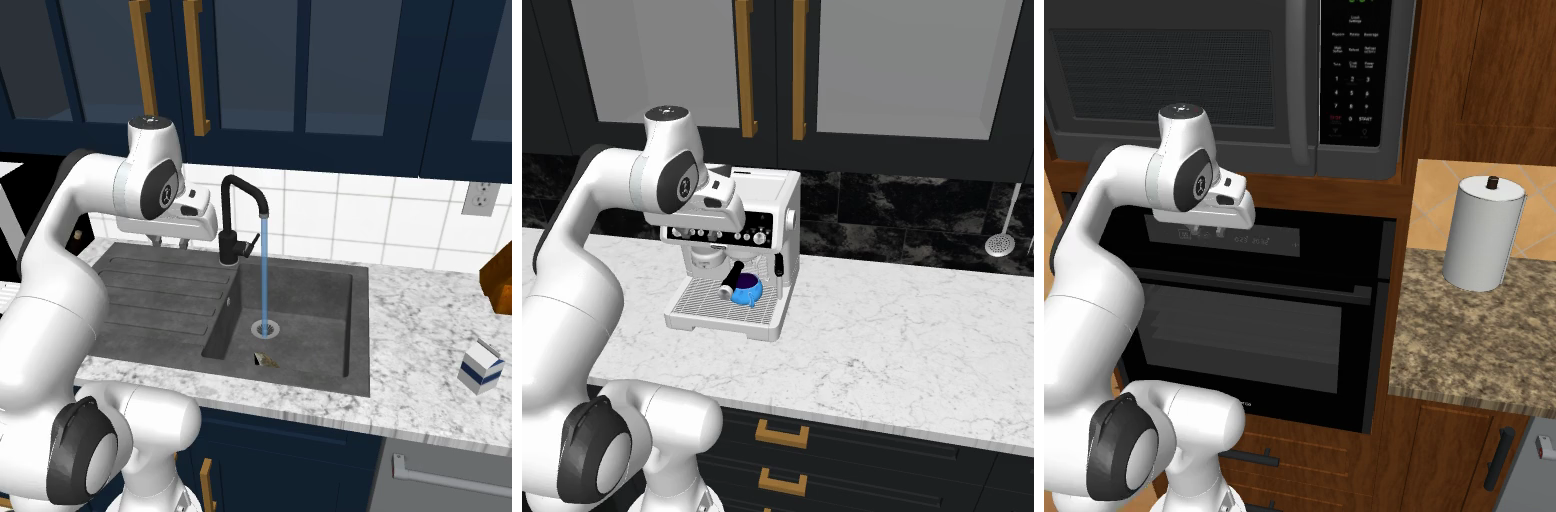}
    \end{subfigure}
    \caption{\textbf{Simulation setups.} We evaluate \ourshort across four widely used robotic manipulation benchmarks spanning diverse task structures and dynamics. These environments cover a range of skills, including object manipulation, tool use, and multi-stage interactions.}
    \label{fig:sim_env_setups}
    \vspace{-15pt}
\end{figure*}

\section{Experiments}
We evaluate \ourshort by comparing autoregressive policies equipped with different action tokenization schemes, as well as non-autoregressive diffusion-based policies. Our experiments assess both downstream policy performance and the impact of key design choices through controlled ablations.

\subsection{Experimental Setup}
Unless otherwise specified, all policies, tokenizers, and evaluation protocols follow the setup described below. We provide more details in Appendix~\ref{appendix:sec:impl_details}.

\subsubsection{Policy Implementation} 
All policies are trained to predict an action chunk of horizon $H_a = 32$ conditioned on the past $H_o = 2$ observations. During execution, we only execute the first $\frac{1}{2}H_a = 16$ actions from each chunk before re-inferring, following standard practice in action chunking.

We evaluate multiple action tokenization schemes within an autoregressive policy framework. We consider per-dimension binning (\bin) and frequency-domain tokenization (\fast). We set \bin vocabulary size to $|\mathcal{V}| = N = 256$, and we use $|\mathcal{V}| = 1024$ for \fast, which are common configurations in prior work. We additionally compare against \textit{Quantized Skill Transformer} (\quest)~\cite{mete2024quest}, a representative learned latent tokenizer. \quest compresses action sequences using a temporal convolution followed by a causal transformer encoder, reducing the temporal horizon from $H_a$ to $H_l$ with a downsampling factor of $4$ (i.e., $H_l = \tfrac{1}{4} H_a$). \quest and \ourshort use the same decoder architecture. \ourshort adopts the same hyperparameters as \quest: a 2-layer transformer encoder with model dimension 256 and head dimension 64, a 4-layer transformer decoder with the same dimensions, latent horizon $H_l = 8$, latent dimension $D_l = 4$, and FSQ levels $[8, 5, 5, 5]$, corresponding to an implicit codebook size $|\mathcal{V}| = 1000$. In addition to autoregressive policies, we include a non-autoregressive baseline based on diffusion policy (\diffp)~\cite{chi2024diffusionpolicy} with a transformer backbone. To isolate the effects of action representation and tokenization, we use the same policy backbone architecture for all methods.

\subsubsection{Evaluation Tasks} 

We conduct comprehensive ablations and analyses, comparing \ourshort against \bin, \fast, \quest, and \diffp across 20+ tasks drawn from 4 widely used simulation benchmarks (Fig.~\ref{fig:sim_env_setups}). Specifically, we evaluate on LIBERO~\cite{liu2023liberobenchmark}, RoboMimic~\cite{robomimic2021}, MetaWorld~\cite{yu2020metaworldbenchmark}, and RoboCasa~\cite{robocasa2024}. For simulation experiments, we evaluate each task across 5 random seeds, with 50 evaluation rollouts per seed, resulting in a total of 250 rollouts per task. We report the mean success rate along with its standard error across rollouts.

\begin{figure}[t]
\captionsetup[subfigure]{justification=raggedright,singlelinecheck=false}
    \centering
    \begin{subfigure}[t]{\linewidth}
        \centering
        \caption{Pick \& Place Ball}
        \includegraphics[width=\linewidth]{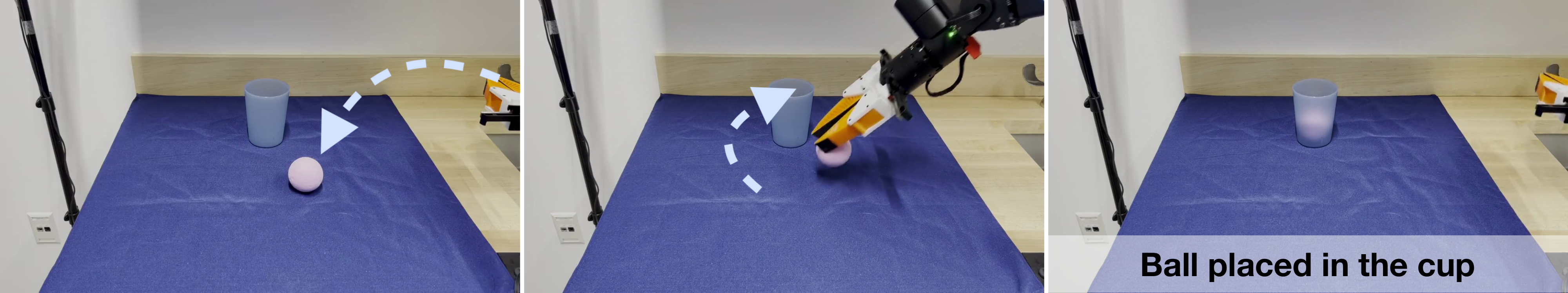}
        \vspace{-15pt}
    \end{subfigure}
    \begin{subfigure}[t]{\linewidth}
        \centering
        \caption{Stack Cups}
        \includegraphics[width=\linewidth]{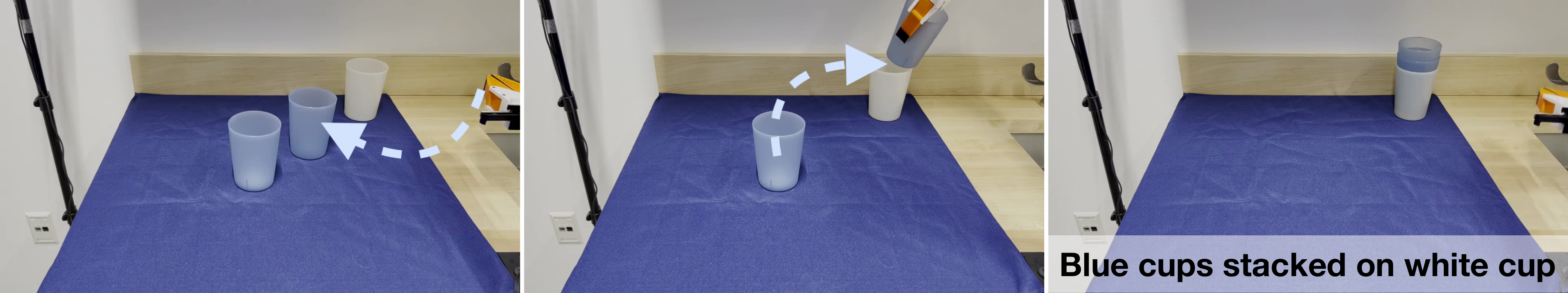}
        \\ \vspace{-2.5pt}
        \begin{tikzpicture}
            \draw[thick,->]
                (0,0) -- (.95\linewidth,0)
                node[right]{\small $t$};
        \end{tikzpicture}
        \vspace{-15pt}
    \end{subfigure}
    \caption{\textbf{Real-world setups.} We validate \ourshort on two tabletop manipulation tasks using a fixed-base robotic arm: (a) \textit{Pick \& Place Ball} and (b) \textit{Stack Cups}. Objects are randomly placed on the table.}
    \label{fig:real_world_setups}
\end{figure}

We additionally validate \ourshort on real-world tabletop manipulation using a fixed-base ARX-5 robotic arm with a single Logitech Webcam for visual observations. We consider two tasks: \textit{Pick \& Place Ball} and \textit{Stack Cups} (Fig.~\ref{fig:real_world_setups}). For each task, we collect 200 human teleoperation demonstrations. The action space is 7D, consisting of end-effector position, orientation, and gripper control. During evaluation, each task is executed for 20 independent trials, and we report task success rates.

\subsection{Simulation Benchmarking}

\begin{table}[t]
\centering
\begin{tabular}{l|cccc}
\toprule
Policy & \multicolumn{1}{c}{LIBERO} & \multicolumn{1}{c}{RoboMimic} & \multicolumn{1}{c}{MetaWorld} & \multicolumn{1}{c}{RoboCasa} \\ \midrule
\diffp       & 36.6 \scriptstderr{0.2} & 67.1 \scriptstderr{1.3} & 19.3 \scriptstderr{1.6} & 54.0 \scriptstderr{1.6} \\
\bin         & 14.4 \scriptstderr{0.6} & 39.5 \scriptstderr{1.2} & 14.5 \scriptstderr{0.7} & 27.7 \scriptstderr{0.9} \\
\fast        & 23.0 \scriptstderr{0.5} & 24.0 \scriptstderr{1.5} &  \phantom{0}7.1 \scriptstderr{0.7} & 13.2 \scriptstderr{1.1} \\
\quest       & 48.2 \scriptstderr{0.6} & 66.9 \scriptstderr{0.8} & 17.9 \scriptstderr{0.9} & 52.3 \scriptstderr{1.9} \\
\midrule
\ourshort[1] & 11.7 \scriptstderr{0.7} & 50.8 \scriptstderr{1.4} & 11.3 \scriptstderr{0.4} & 47.7 \scriptstderr{1.3} \\
\ourshort[2] & 39.8 \scriptstderr{0.5} & 52.5 \scriptstderr{1.2} & 16.4 \scriptstderr{0.3} & 50.3 \scriptstderr{0.8} \\
\ourshort[4] & 46.4 \scriptstderr{0.6} & 65.3 \scriptstderr{0.9} & 19.5 \scriptstderr{0.8} & 51.7 \scriptstderr{1.0} \\
\ourshort[8] & 56.3 \scriptstderr{1.0} & 73.1 \scriptstderr{0.5} & 24.4 \scriptstderr{0.3} & 54.6 \scriptstderr{1.1} \\
\bottomrule
\end{tabular}
\caption{\textbf{Simulation benchmarking} across four manipulation benchmarks. \ourshort consistently outperforms prior action tokenization schemes and exhibits monotonic performance improvements as the number of decoded tokens increases. \ourshort[K] denotes detokenization using the first $K$ tokens. Results report mean success rates with standard error across 5 seeds and 50 evaluation rollouts per seed per task. Complete results in Appendix~\ref{appendix:sec:sim_benchmarking}.}
\label{table:simulation_benchmarking}
\end{table}

Table~\ref{table:simulation_benchmarking} reports performance across four simulation benchmarks. \bin performs poorly across all benchmarks, as it produces excessively long token sequences and thus violates \propertyI. \fast achieves compact representations but suffers from invalid or non-decodable token sequences, violating \propertyII and leading to unstable policy behavior. Notably, both methods exhibit high reconstruction fidelity, confirming that reconstruction error alone is not predictive of downstream policy performance. \quest provides a substantially stronger baseline by leveraging quantized latent actions. However, its latent token space lacks an ordering, violating \propertyIII, hence its autoregressive modeling does not benefit from inductive biases from causal token ordering aligned with next-token prediction.

\ourshort consistently outperforms prior action tokenization schemes and matches or exceeds the strongest baselines, while additionally enabling prefix-based decoding that is unavailable to existing methods. We denote \ourshort[K] as executing action chunks reconstructed from the first $K$ tokens, i.e., detokenizing the prefix $T_{1:K}$ with $K \leq H_l$. \ourshort exhibits a clear and consistent monotonic performance trend as the number of autoregressive steps increases. As additional tokens are generated, performance improves steadily: \ourshort[4] closes much of the gap to \quest and \diffp, while \ourshort[8] achieves the best performance across all benchmarks. This enables an anytime trade-off between computation and performance: policies may terminate autoregressive generation early when latency constraints are tight, or generate longer sequences for improved performance.

\subsection{Ablation and Analysis}
\subsubsection{\textbf{Compression Rate and Inference Latency}}

\begin{table}[t]
\centering
\setlength{\tabcolsep}{4.0pt}
\begin{tabular}{l | rr | rr | rr | rr}
\toprule
& \multicolumn{2}{c|}{LIBERO} & \multicolumn{2}{c|}{RoboMimic} & \multicolumn{2}{c|}{MetaWorld} & \multicolumn{2}{c}{RoboCasa} \\
Policy & \#Tok. & Lat. & \#Tok. & Lat. & \#Tok. & Lat. & \#Tok. & Lat. \\
\midrule
\diffp   & $\times$ & 42.0 & $\times$ & 38.1 & $\times$ & 37.7 & $\times$ & 35.3 \\
\bin     & 224 & 517.2 & 224 & 509.5 & 128 & 306.6 & 384 & 888.3 \\
\fast    & 44.2 & 114.4 & 53.1 & 142.0 & 49.8 & 129.7 & 69.7 & 166.1 \\
\quest   & 8 & 27.1  & 8 & 29.6  & 8 & 31.4  & 8 & 30.2 \\
\midrule
\ourshort[1] & 1 & 10.5  & 1 & 11.3  & 1 & 15.5  & 1 & 13.5 \\
\ourshort[2] & 2 & 13.2  & 2 & 15.3  & 2 & 17.9  & 2 & 15.8 \\
\ourshort[4] & 4 & 17.4  & 4 & 18.4  & 4 & 22.1  & 4 & 19.8 \\
\ourshort[8] & 8 & 27.4  & 8 & 29.9  & 8 & 31.3  & 8 & 30.0 \\
\bottomrule
\end{tabular}
\caption{\textbf{Token count and inference latency.} Comparison of action token counts (\#Tok.$\downarrow$) and policy inference times (Lat.$\downarrow$) across various benchmarks. For \fast, which generates variable-length sequences, we report the mean token count. \ourshort[K] denotes detokenization using the first $K$ tokens. Policy latency is measured in milliseconds (ms) per inference on one NVIDIA A100.}
\label{table:num_tokens}
\end{table}

Table~\ref{table:num_tokens} compares action compression rates and inference latency across methods. \bin produces extremely long token sequences, resulting in prohibitively high inference latency, while \fast achieves only moderate compression. \quest compresses each action chunk into a fixed-length token sequence, yielding significantly lower inference latency. However, its fixed decoding length limits flexibility. \ourshort enables a smooth and controllable trade-off between compression rate, inference latency, and policy performance. With full decoding, \ourshort and \quest have the same amount of compute per inference.

\subsubsection{\textbf{How Token Space Ordering (\propertyIII) Matters?}}
\label{sec:token_ordering_ablation}

\begin{table}[t]
\centering
\begin{tabular}{l|cccc}
\toprule
Policy & \multicolumn{1}{c}{LIBERO} & \multicolumn{1}{c}{RoboMimic} & \multicolumn{1}{c}{MetaWorld} & \multicolumn{1}{c}{RoboCasa} \\ \midrule
\quest       & 48.2 \scriptstderr{0.6} & 66.9 \scriptstderr{0.8} & 17.9 \scriptstderr{0.9} & 52.3 \scriptstderr{1.9} \\
\midrule
\ourshort[1] & 11.7 \scriptstderr{0.7} & 50.8 \scriptstderr{1.4} & 11.3 \scriptstderr{0.4} & 47.7 \scriptstderr{1.3} \\
\ourshort[2] & 39.8 \scriptstderr{0.5} & 52.5 \scriptstderr{1.2} & 16.4 \scriptstderr{0.3} & 50.3 \scriptstderr{0.8} \\
\ourshort[4] & 46.4 \scriptstderr{0.6} & 65.3 \scriptstderr{0.9} & 19.5 \scriptstderr{0.8} & 51.7 \scriptstderr{1.0} \\
\ourshort[8] & 56.3 \scriptstderr{1.0} & 73.1 \scriptstderr{0.5} & 24.4 \scriptstderr{0.3} & 54.6 \scriptstderr{1.1} \\
\midrule
\ourshort[\times] & 35.2 \scriptstderr{0.7} & 61.1 \scriptstderr{1.2} & 17.6 \scriptstderr{0.5} & 48.5 \scriptstderr{1.6} \\
\bottomrule
\end{tabular}
\caption{\textbf{\ourshort without ordering underperforms.} Simulation benchmarking across four manipulation benchmarks. \ourshort[K] denotes detokenization using the first $K$ tokens, while \ourshort[\times] denotes tokenizer training without nested dropout. Results report mean success rates with standard error across 5 seeds and 50 evaluation rollouts per seed per task.}
\label{table:token_ordering_ablation}
\end{table}

Table~\ref{table:token_ordering_ablation} studies the role of token space ordering by comparing \ourshort trained with and without ordering-inducing mechanisms, i.e., nested dropout, which enforces a left-to-right priority structure over tokens during training. The variant \ourshort[\times] disables nested dropout, resulting in an unordered token space, while all other architectural and training settings are kept identical. 

Across all benchmarks, removing token ordering leads to a consistent performance degradation. \ourshort[\times]'s performance is significantly worse than \ourshort[4] and \ourshort[8], and in some cases falls below \quest. This indicates that the structure of the token space plays a critical role in effective autoregressive policy learning: by aligning the token space with next-token prediction, ordering introduces a favorable inductive bias that facilitates both learning and inference.

\subsubsection{\textbf{How Action (\texorpdfstring{$H_a$}{Ha}) and Latent (\texorpdfstring{$H_l$}{Hl}) Horizon Matter?}}

\begin{figure}[t]
    \centering
    \begin{subfigure}[t]{0.45\linewidth}
        \centering
        \includegraphics[width=\linewidth]{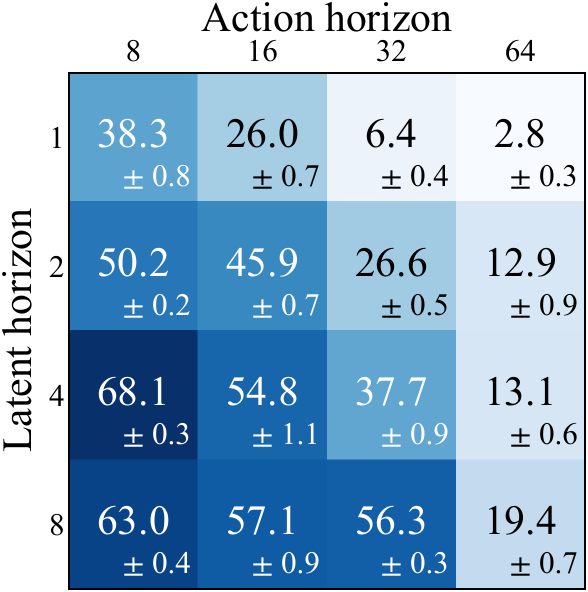}
        \caption{Execute $\frac{1}{2}H_a$ actions}
        \label{fig:Ha_Hl_ablation_half_H_a}
    \end{subfigure}
    \hspace{5pt}
    \begin{subfigure}[t]{0.45\linewidth}
        \centering
        \includegraphics[width=\linewidth]{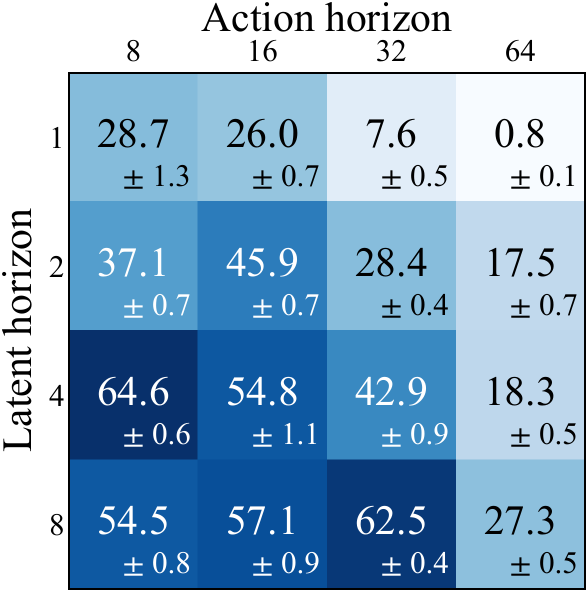}
        \caption{Execute a fixed 8 actions}
        \label{fig:Ha_Hl_ablation_fixed_8}
    \end{subfigure}
    \caption{\textbf{Effect of action and token horizons.} Performance of \ourshort[H_l] on LIBERO as a function of action horizon $H_a$ (rows) and token horizon $H_l$ (columns). Results report mean success rates with standard error across 5 seeds and 50 evaluation rollouts per seed per task.}
    \label{fig:Ha_Hl_ablation}
\end{figure}

Table~\ref{fig:Ha_Hl_ablation} analyzes the interaction between action horizon $H_a$ and latent token horizon $H_l$ for \ourshort on LIBERO. The latent horizon $H_l$ is a training-time hyperparameter that determines the number of register tokens. We train separate models for all combinations of $H_a \in \{8,16,32,64\}$ and $H_l \in \{1,2,4,8\}$. To disentangle modeling effects from execution effects, we report two execution regimes: executing $\tfrac{1}{2}H_a$ actions before re-inference (Table~\ref{fig:Ha_Hl_ablation_half_H_a}), which reflects practical receding-horizon control, and executing a fixed 8 actions for all $H_a$ (Table~\ref{fig:Ha_Hl_ablation_fixed_8}) as a controlled diagnostic.

Under the practical execution regime (Table~\ref{fig:Ha_Hl_ablation_half_H_a}), performance degrades monotonically with increasing $H_a$ for a fixed $H_l$, reflecting the growing difficulty of long-horizon prediction under limited latent capacity. Increasing $H_l$ consistently mitigates this effect, indicating that additional register tokens enable finer-grained temporal encoding. However, when $H_a \leq H_l$, the information bottleneck largely disappears, yielding diminishing returns; prior work suggests that moderate bottlenecks are beneficial for learning~\cite{dieleman2025latents, he2024bridgsim2realinfobottleneck, li2025basicsletdenoisinggenerative}, explaining the observed saturation for short horizons such as $H_a=8$. The fixed-execution setting (Table~\ref{fig:Ha_Hl_ablation_fixed_8}) reveals a complementary trend. For a fixed $H_l$, performance becomes non-monotonic in $H_a$: moderate horizons improve performance by stabilizing early actions~\cite{zhang2025actionchunkingexploratorydata}, while excessively long horizons degrade performance due to the difficulty of compressing long futures into a limited number of tokens.

Together, these results highlight the trade-off between temporal lookahead and latent capacity. Predicting beyond the execution horizon can improve robustness and consistency, but only when the tokenizer can faithfully represent the future. Although the fixed-step execution regime is not intended to reflect deployment, it provides a useful diagnostic when interpreted alongside the receding-horizon setting. These findings motivate our default choice of $H_a=32$ and $H_l=8$, which balances long-horizon expressivity, compression, and execution stability.

\subsubsection{\textbf{How Codebook Size Matters?}}

\begin{table}[t]
\centering
\setlength{\tabcolsep}{4.0pt}
\begin{tabular}{c|ccccc}
\toprule
FSQ Levels    & [8,6,5] & [8,8,8] & [8,5,5,5] & [8,8,6,5] & [7,5,5,5,5] \\
Induced $|\mathcal{V}|$ & 240 & 512 & 1000 & 1920 & 4375 \\
\midrule
LIBERO       & 29.2 \scriptstderr{0.8} & 53.5 \scriptstderr{1.2} & 56.3 \scriptstderr{1.0} & 54.6 \scriptstderr{1.1} & 46.9 \scriptstderr{0.6} \\
\bottomrule
\end{tabular}
\caption{\textbf{Effect of codebook size.} Performance of \ourshort on LIBERO under varying FSQ codebook sizes. Results are relatively insensitive to codebook size once moderate capacity is reached, while excessively large codebooks degrade downstream autoregressive learning. Results report mean success rate with standard error across 5 seeds and 50 evaluation rollouts per seed per task.}
\label{tab:codebook_size_ablation}
\vspace{-10pt}
\end{table}

Table~\ref{tab:codebook_size_ablation} examines the impact of discrete codebook size $|\mathcal{V}|$, controlled via FSQ level configurations. We vary $|\mathcal{V}|$ from $2^8$ to $2^{12}$ while keeping all other architectural and training settings fixed. Performance improves substantially as the codebook capacity increases from very small to moderate, after which it saturates. However, further enlarging the codebook leads to a clear performance drop. We attribute this degradation to reduced \textit{modelability} for downstream autoregressive policies: larger codebooks increase token entropy and sparsity, making next-token prediction more difficult despite improved reconstruction fidelity.

\subsection{Real-world Results}


\begin{table}[t]
\centering
\setlength{\tabcolsep}{5.0pt}
\begin{minipage}[t]{0.5\linewidth}
\centering
\begin{tabular}{l|cc}
\toprule
Policy & P\&P Ball & Stack Cups \\ 
\midrule
\diffp       & 14 / 20 & 11 / 20 \\
\bin         & \phantom{0}4 / 20 & \phantom{0}8 / 20 \\
\fast        & \phantom{0}8 / 20 & \phantom{0}6 / 20 \\
\quest       & 11 / 20 & \phantom{0}8 / 20 \\
\midrule
\ourshort[1] & \phantom{0}7 / 20 & \phantom{0}3 / 20 \\
\ourshort[2] & 11 / 20 & \phantom{0}9 / 20 \\
\ourshort[4] & 13 / 20 & 12 / 20 \\
\ourshort[8] & 16 / 20 & 16 / 20 \\
\bottomrule
\end{tabular}
\end{minipage}
\hfill
\begin{minipage}[t]{0.48\linewidth}
\vspace{-51.5pt}
\captionof{table}{\textbf{Real-world results} on two manipulation tasks. \ourshort consistently outperforms others, and performance improves as the number of decoded tokens increases. \ourshort[K] denotes detokenization using the first $K$ tokens. We report mean success rates over 20 evaluation rollouts per task.}
\label{table:realworld_benchmarking}
\end{minipage}
\end{table}

Table~\ref{table:realworld_benchmarking} reports real-world performance on two tabletop manipulation tasks. The results closely mirror trends observed in simulation, validating that the benefits of ordered, prefix-decodable action tokens transfer to real-world robotic control. \bin performs poorly primarily due to excessive latency induced by long token sequences, which degrades closed-loop responsiveness. \fast, despite its compact tokenization, fails to decode consistently and exhibits unstable, overly aggressive behavior, preventing reliable task execution. \quest improves over these baselines but remains limited by its unstructured latent representation. 

\ourshort consistently achieves the highest success rates across both tasks, with performance improving monotonically as the number of decoded tokens increases. Beyond success rates, we observe clear qualitative differences in trajectory execution. \ourshort produces noticeably smoother motions, with smoothness improving as more tokens are decoded. A common failure mode for \ourshort[<4] is insufficient execution precision: the robot often reaches configurations that are visually close to success but fails to complete fine-grained insertions (e.g., placing the ball fully into the cup). This behavior indicates that early tokens capture coarse, global action structure, while later tokens encode fine-grained corrective details necessary for precise manipulation, directly supporting the design intent of ordered tokenization.

\section{Discussion and Limitations}

This work introduces \ourshort, an action tokenization framework for autoregressive policies that emphasizes ordered, prefix-decodable action representations. While our results demonstrate strong performance and flexibility, several broader implications and open challenges remain.

Recent VLA systems increasingly combine discrete reasoning with continuous control by integrating multiple policy components. For example, the BEHAVIOR-1K~\cite{li2024behavior1k} winning system~\cite{larchenko2025taskadaptationvla} employs \fast as an auxiliary discrete action representation alongside continuous flow-based experts, highlighting an emerging paradigm in which action tokenization complements rather than replaces continuous policies. In this context, \ourshort offers a principled alternative: its left-to-right ordered and prefix-decodable token space supports autoregressive reasoning over actions while remaining compatible with continuous decoders such as diffusion or flow models. This makes \ourshort a natural auxiliary supervision signal, planning interface, or intermediate abstraction for future VLA pipelines.

A key capability enabled by \ourshort is prefix-based detokenization, which allows actions to be decoded from variable-length token prefixes and provides an anytime trade-off between computation and action fidelity. In this work, however, the autoregressive depth is fixed at deployment time. From an information-theoretic perspective, this is suboptimal: the number of tokens required to represent an action chunk $a_{1:H_a}$ should depend on its intrinsic complexity and required precision. Simple behaviors may admit compact representations, while complex, contact-rich interactions may require deeper autoregressive refinement. Estimating action complexity online and deciding when additional tokens meaningfully reduce uncertainty remains an open problem. We view adaptive autoregressive depth as a natural and important direction for future work, enabled precisely by \ourshort’s ordered and prefix-decodable structure.

\clearpage
\section*{Acknowledgments}
The computations in this paper were run on the FASRC cluster supported by the FAS Division of Science Research Computing Group at Harvard University. We thank the members of the Embodied Minds Lab at Harvard for insightful discussions, constructive feedback during early stages of this work, and assistance with manuscript proofreading.

\bibliographystyle{plainnat}
\bibliography{references}

\clearpage
\appendix
\section{Appendix}

\subsection{Implementation Details}
\label{appendix:sec:impl_details}

This section provides comprehensive implementation details for all policies, tokenizers, optimization settings, and evaluation protocols referenced in the main paper.

All policies are trained to predict a contiguous action chunk of horizon $H_a = 32$, conditioned on the most recent $H_o = 2$ observations. During deployment, policies operate in a receding-horizon manner: only the first $\tfrac{1}{2}H_a = 16$ actions of each predicted chunk are executed before re-inference. This execution strategy balances temporal consistency and responsiveness, and is used consistently across all methods unless stated otherwise.

We compare multiple action tokenization schemes within the same autoregressive policy framework:
\begin{itemize}
    \item \bin~\cite{rt12022arxiv, rt22023arxiv, kim24openvla}: Each action dimension is discretized into $N=256$ uniform bins, yielding a token sequence whose length scales with $H_a \times D_a$.
    \item \fast~\cite{pertsch2025fast}: We use a vocabulary size of $|\mathcal{V}|=1024$, following standard configurations in prior work.
    \item \quest~\cite{mete2024quest}: Action chunks are compressed using a temporal convolution followed by a causal transformer encoder, reducing the temporal horizon from $H_a$ to $H_l = \tfrac{1}{4}H_a = 8$.
    \item \ourshort: For fair comparison, \ourshort adopts the same decoder architecture and latent dimensionality as \quest. The tokenizer encoder is a 2-layer transformer with model dimension 256 and head dimension 64. The latent representation has horizon $H_l=8$ and latent dimension $D_l=4$, discretized using finite scalar quantization (FSQ) with levels $[8,5,5,5]$, corresponding to an implicit codebook size of approximately 1000.
\end{itemize}

All autoregressive policies share the same backbone architecture: a transformer decoder with 4 layers, model dimension 256, and head dimension 64. The decoder predicts discrete action tokens autoregressively, using teacher forcing during training and fully autoregressive rollout during inference. Using a shared policy backbone isolates the effect of action representation from policy capacity. In addition to autoregressive policies, we include a diffusion-based baseline (\diffp)~\cite{chi2024diffusionpolicy}. The diffusion policy uses exactly the same 4-layer transformer backbone as the autoregressive models, ensuring that performance differences arise from the action representation and inference paradigm rather than architectural capacity. We employ a 10-step Denoising Diffusion Implicit Models (DDIM)~\cite{song2021ddim} sampling schedule for \diffp. All models are trained using AdamW with identical optimization settings: a constant learning rate of \texttt{5e-5} for tokenizers and policy networks, and \texttt{1e-5} for observation encoders, with no weight decay.

We evaluate and analyze policies on four widely used simulation benchmarks:
\begin{itemize}
    \item LIBERO~\cite{liu2023liberobenchmark}: \textit{libero10}; 50 demonstrations per task; action dimension $D_a = 7$.
    \item RoboMimic~\cite{robomimic2021}: \textit{lift, square, can}; 200 multi-human (mh) demonstrations per task; action dimension $D_a = 7$.
    \item MetaWorld~\cite{yu2020metaworldbenchmark}: \textit{box close, coffee pull, disassemble, stick pull}; 50 demonstrations per task; action dimension $D_a = 4$.
    \item RoboCasa~\cite{robocasa2024}: \textit{close drawer, coffee press button, turn off microwave, turn off sink faucet}; 50 human demonstrations and 150 machine-generated demonstrations per task; action dimension $D_a = 12$.
\end{itemize}
For each task, we evaluate 5 random seeds with 50 rollouts per seed, resulting in 250 evaluation episodes per task. Performance is reported as the mean task success rate with standard error across rollouts.

\subsection{The Structural Mismatch of \fast}
\label{appendix:sec:fast_decoding_error}

A critical limitation of the \fast tokenizer arises from the fundamental structural conflict between the probabilistic, variable-length nature of Byte Pair Encoding (BPE)~\cite{gage1994bpe} and the strict, fixed-dimensional requirements of robotic control.

\subsubsection{Mechanism of \fast.}
\fast operates by applying a Discrete Cosine Transform (DCT)~\cite{ahmed1974dct, cooley1965fourierseries} to action chunks, pruning low-magnitude high-frequency components, and flattening the remaining coefficients into a 1D integer sequence. A BPE tokenizer is then trained to compress this sequence. While this effectively separates coarse structure from fine detail, it introduces a critical dependency between the token sequence length and the action chunk topology.

\subsubsection{Variable Expansion vs. Fixed Topology}
In standard large language models, the decoding process is agnostic to the exact number of characters produced; a token representing \texttt{apple} (5 bytes) is structurally valid in the same context as \texttt{a} (1 byte). However, the \fast tokenizer maps discrete tokens to variable-length sequences of DCT coefficients. Let a generated token sequence be $ T_{1:H_l} = [T_1, T_2, \dots, T_{H_l}] $. Each token $T_i$ expands into a sequence of integers $s_i$ of length $|s_i|$. The total sequence of coefficients $S$ is the concatenation of these expansions:
\[
S = s_1 \oplus s_2 \oplus \dots \oplus s_k, \quad \text{where } |S| = \sum_{i=1}^k |s_i|.
\]
The robot controller, however, strictly requires a control chunk of dimensions $H_a \times D_a$ (time horizon $\times$ action dimension), necessitating a fixed total coefficient count $N_{\text{target}} = T \times D$. 

\subsubsection{The Decoding Failure}
Because the policy is autoregressive and probabilistic, it generates tokens based on likelihood rather than structural constraints. There is no guarantee that the generated sequence $T_{1:H_l}$ will satisfy $|S| = N_{\text{target}}$. When $|S| \neq N_{\text{target}}$, the reshaping operation into $(H_a, D_a)$ becomes mathematically impossible, raising the ``undecodable'' error (e.g., \texttt{ValueError: cannot reshape array}).

\subsubsection{The ``Spectral Shift'' Catastrophe}
Naive solutions, such as padding or truncating $S$ to match $N_{\text{target}}$, are catastrophic due to the use of the discrete cosine transform (DCT)~\cite{ahmed1974dct, cooley1965fourierseries}. The sequence $S$ is an ordered flattening of frequency coefficients. If a token generating 3 coefficients is replaced by a token generating 2 coefficients (a ``missing'' coefficient at index $j$), every subsequent coefficient at indices $k > j$ shifts position. In the frequency domain, this shift is semantically destructive, for example, coefficients governing joint $J$ may drift into the slots for joint $J+1$. Consequently, the undecodable state acts as a necessary safety assertion. It is preferable to halt execution (outputting a no-op) than to reshape a corrupted coefficient sequence that would result in unpredictable and potentially dangerous physical motion.

\subsection{\texorpdfstring{\ourshort}{} Detokenization \texorpdfstring{$\mathcal{T}^{-1}$}{invT}}
\label{appendix:sec:our_tok_dec}

Similar to \cite{zhao2023actpolicy}, the single-pass decoder is implemented as a transformer decoder consisting of alternating self-attention and cross-attention layers. The decoder cross-attends from a fixed set of sinusoidal positional embeddings to the discrete action tokens produced by \ourshort. The final decoder embeddings are projected back into the continuous action space, yielding a reconstructed action chunk of shape $H_a \times D_a$. The tokenizer and decoder are trained end-to-end using a reconstruction objective, specifically mean squared error (MSE) between the original and reconstructed action chunks.

When the latent bottleneck is small, training the decoder with a simple reconstruction loss can lead to degraded reconstruction quality, as the decoder must recover long-horizon action sequences from severely compressed representations~\cite{dieleman2025latents, lipman2023flowmatchinggenerativemodeling, lipman2024flowmatchingguidecode}. To address this limitation, \ourshort can employ a rectified flow decoder conditioned on the quantized register latents. Concretely, we construct partially noised action sequences
\begin{equation*}
a_{1:H_a}^t = (1 - t)\, a_{1:H_a}^0 + t\, \epsilon,
\end{equation*}
where $a_{1:H_a}^0$ denotes the clean action chunk, $t \in [0,1]$ is a randomly sampled time step, and $\epsilon \sim \mathcal{N}(0, I)$ is Gaussian noise. The flow decoder receives the concatenation of the noised actions and the quantized register tokens $\textit{Quant}(z_{1:H_l})$ and is trained to predict the flow
\begin{equation*}
v = \epsilon - a_{1:H_a}^0.
\end{equation*}
We minimize the rectified flow objective $\lVert \hat{v} - v \rVert^2$, where
\begin{equation*}
\hat{v} = \textit{Dec}\!\left(\textit{Quant}(z_{1:H_l}) \oplus a_{1:H_a}^t\right),
\end{equation*}
following prior work on flow-based generative modeling~\cite{liu2022flowstraightfastlearning, bachmann2025flextok}.



\subsection{Simulation Benchmarking}
\label{appendix:sec:sim_benchmarking}

We provide full results of simulation experiments in Table~\ref{appendix:table:all_sim_results}.

\begin{table*}[t]
\centering
\scriptsize
\setlength{\tabcolsep}{4.75pt}
\begin{tabular}{l|rr|cccccccccc|c}
\toprule
\multicolumn{14}{c}{\textbf{LIBERO}} \\
\midrule
Policy & 
\#Tok. & 
\makecell{Inf.\\Lat.} & 
\makecell{Soup/Sauce\\Basket} & 
\makecell{Cheese/Butter\\Basket} & 
\makecell{Soup/Cheese\\Basket} & 
\makecell{Two Moka\\Pots} & 
\makecell{Stove \&\\Moka} & 
\makecell{Bowl to\\Drawer} & 
\makecell{Mugs on\\Plates} & 
\makecell{Book to\\Caddy} & 
\makecell{Mug \&\\Pudding} & 
\makecell{Mug to\\Micro} & 
\makecell{Avg.} \\ 
\midrule
\diffp & $\times$ & 42.0 & 26.0 \tinystderr{3.0} & 18.8 \tinystderr{1.4} & 24.8 \tinystderr{1.9} & 52.4 \tinystderr{2.7} & 56.8 \tinystderr{3.4} & 62.8 \tinystderr{2.1} & 20.0 \tinystderr{1.7} & 18.4 \tinystderr{1.9} & 29.6 \tinystderr{2.6} & 56.0 \tinystderr{3.1} & 36.6 \tinystderr{0.2} \\
\bin & 224 & 517.2 & \phantom{0}1.6 \tinystderr{0.7} & \phantom{0}3.6 \tinystderr{0.7} & \phantom{0}4.0 \tinystderr{1.7} & \phantom{0}8.8 \tinystderr{2.0} & 24.0 \tinystderr{1.1} & 46.0 \tinystderr{3.4} & \phantom{0}2.8 \tinystderr{1.5} & 31.2 \tinystderr{2.5} & \phantom{0}6.8 \tinystderr{0.8} & 15.6 \tinystderr{2.4} & 14.4 \tinystderr{0.6} \\
\fast & 44.2 & 114.4 & 14.8 \tinystderr{1.6} & \phantom{0}6.4 \tinystderr{0.7} & \phantom{0}1.6 \tinystderr{0.7} & 33.6 \tinystderr{4.4} & 52.8 \tinystderr{1.4} & 50.4 \tinystderr{5.0} & 16.0 \tinystderr{1.7} & 28.4 \tinystderr{1.7} & 22.0 \tinystderr{3.5} & \phantom{0}4.4 \tinystderr{1.3} & 23.0 \tinystderr{0.5} \\
\quest & 8 & 27.1 & 22.4 \tinystderr{2.8} & 16.0 \tinystderr{2.8} & 31.6 \tinystderr{2.7} & 47.6 \tinystderr{2.0} & 79.6 \tinystderr{1.9} & 88.0 \tinystderr{1.4} & 20.8 \tinystderr{2.8} & 65.6 \tinystderr{4.8} & \highlightcell 35.6 \tinystderr{1.3} & 74.8 \tinystderr{3.7} & 48.2 \tinystderr{0.6} \\
\ourshort[1] & 1 & 10.5 & \phantom{0}2.4 \tinystderr{0.7} & \phantom{0}1.6 \tinystderr{0.7} & \phantom{0}1.6 \tinystderr{0.7} & \phantom{0}2.8 \tinystderr{0.8} & 23.6 \tinystderr{1.2} & 26.0 \tinystderr{2.9} & \phantom{0}0.8 \tinystderr{0.5} & 26.8 \tinystderr{3.4} & \phantom{0}3.6 \tinystderr{1.0} & 28.0 \tinystderr{1.7} & 11.7 \tinystderr{0.7} \\
\ourshort[2] & 2 & 13.2 & 15.2 \tinystderr{3.1} & 16.4 \tinystderr{1.3} & 25.2 \tinystderr{2.1} & 39.2 \tinystderr{1.5} & 59.2 \tinystderr{2.4} & 69.6 \tinystderr{4.3} & 14.0 \tinystderr{1.4} & 81.2 \tinystderr{1.9} & 13.6 \tinystderr{3.0} & 64.8 \tinystderr{2.9} & 39.8 \tinystderr{0.5} \\
\ourshort[4] & 4 & 17.4 & 14.8 \tinystderr{1.4} & 16.4 \tinystderr{1.7} & 32.4 \tinystderr{2.2} & 57.2 \tinystderr{3.2} & 68.8 \tinystderr{3.1} & 78.4 \tinystderr{2.7} & 24.4 \tinystderr{1.5} & \highlightcell 86.0 \tinystderr{2.1} & 14.8 \tinystderr{4.8} & 70.8 \tinystderr{2.2} & 46.4 \tinystderr{0.6} \\
\ourshort[8] & 8 & 27.4 & \highlightcell 26.8 \tinystderr{3.2} & \highlightcell 35.6 \tinystderr{2.6} & \highlightcell 51.6 \tinystderr{2.2} & \highlightcell 61.2 \tinystderr{4.3} & \highlightcell 87.6 \tinystderr{1.2} & \highlightcell 91.2 \tinystderr{1.0} & \highlightcell 31.2 \tinystderr{2.7} & 70.8 \tinystderr{4.5} & 32.0 \tinystderr{2.8} & \highlightcell 75.2 \tinystderr{3.8} & \highlightcell 56.3 \tinystderr{1.0} \\
\ourshort[\times] & 8 & 27.4 & \phantom{0}5.6 \tinystderr{1.6} & \phantom{0}5.6 \tinystderr{0.7} & 21.2 \tinystderr{4.1} & 33.6 \tinystderr{2.4} & 65.6 \tinystderr{1.2} & 81.2 \tinystderr{3.4} & \phantom{0}6.0 \tinystderr{1.1} & 73.6 \tinystderr{4.0} & \phantom{0}3.2 \tinystderr{1.6} & 56.0 \tinystderr{2.2} & 35.2 \tinystderr{0.7} \\
\toprule
\multicolumn{14}{c}{\textbf{RoboMimic}} \\
\midrule
Policy & 
\#Tok. & 
\makecell{Inf.\\Lat.} & 
\makecell{Lift} & 
\makecell{Square} & 
\makecell{Can} & 
\multicolumn{7}{c|}{} & 
\makecell{Avg.} \\ 
\midrule
\diffp            & $\times$ & 38.1  & \highlightcell 99.6 \tinystderr{0.4} & 24.0 \tinystderr{1.8} & 77.6 \tinystderr{2.6} & \multicolumn{7}{c|}{} & 67.1 \tinystderr{1.3} \\
\bin              & 224 & 509.5 & 86.0 \tinystderr{1.3} &  \phantom{0}1.2 \tinystderr{0.8} & 31.2 \tinystderr{2.4} & \multicolumn{7}{c|}{} & 39.5 \tinystderr{1.2} \\
\fast             & 53.1 & 142.0 & 53.6 \tinystderr{3.0} &  \phantom{0}0.4 \tinystderr{0.4} & 18.0 \tinystderr{3.2} & \multicolumn{7}{c|}{} & 24.0 \tinystderr{1.5} \\
\quest            & 8   & 29.6  & 98.8 \tinystderr{0.5} & 29.2 \tinystderr{4.8} & 72.8 \tinystderr{3.0} & \multicolumn{7}{c|}{} & 66.9 \tinystderr{0.8} \\
\ourshort[1]      & 1   & 11.3  & 89.6 \tinystderr{1.5} &  \phantom{0}6.4 \tinystderr{1.2} & 56.4 \tinystderr{3.4} & \multicolumn{7}{c|}{} & 50.8 \tinystderr{1.4} \\
\ourshort[2]      & 2   & 15.3  & 86.6 \tinystderr{1.6} & 11.2 \tinystderr{0.8} & 59.6 \tinystderr{2.8} & \multicolumn{7}{c|}{} & 52.5 \tinystderr{1.2} \\
\ourshort[4]      & 4   & 18.4  & 99.2 \tinystderr{0.5} & 23.6 \tinystderr{1.6} & 73.2 \tinystderr{2.7} & \multicolumn{7}{c|}{} & 65.3 \tinystderr{0.9} \\
\ourshort[8]      & 8   & 29.9  & 99.2 \tinystderr{0.5} & \highlightcell 39.2 \tinystderr{2.4} & \highlightcell 80.8 \tinystderr{2.3} & \multicolumn{7}{c|}{} & \highlightcell 73.1 \tinystderr{0.5} \\
\ourshort[\times] & 8   & 29.2  & 96.8 \tinystderr{1.0} & 16.0 \tinystderr{4.2} & 70.4 \tinystderr{4.9} & \multicolumn{7}{c|}{} & 61.1 \tinystderr{1.2} \\
\toprule
\multicolumn{14}{c}{\textbf{MetaWorld}} \\
\midrule
Policy & 
\#Tok. & 
\makecell{Inf.\\Lat.} & 
\makecell{Box\\Close} & 
\makecell{Coffee\\Pull} & 
\makecell{Disassemble} & 
\makecell{Stick\\Pull} & 
\multicolumn{6}{c|}{} & 
\makecell{Avg.} \\ 
\midrule
\diffp       & $\times$ & 37.7  & 21.2 \tinystderr{4.6} & 27.6 \tinystderr{1.3} & \highlightcell 23.2 \tinystderr{1.0} & \phantom{0}5.2 \tinystderr{0.5} & \multicolumn{6}{c|}{} & 19.3 \tinystderr{1.6} \\
\bin         & 128  & 306.6 &  \phantom{0}9.6 \tinystderr{2.6} & 24.4 \tinystderr{0.7} & 20.8 \tinystderr{1.6} & \phantom{0}3.2 \tinystderr{0.5} & \multicolumn{6}{c|}{} & 14.5 \tinystderr{0.7} \\
\fast        & 49.8 & 129.7 &  \phantom{0}0.0 \tinystderr{0.0} & 16.4 \tinystderr{2.0} & 10.4 \tinystderr{2.6} & \phantom{0}1.6 \tinystderr{0.7} & \multicolumn{6}{c|}{} &  \phantom{0}7.1 \tinystderr{0.7} \\
\quest       & 8    & 31.4  & 12.4 \tinystderr{2.0} & \highlightcell 28.4 \tinystderr{1.8} & \highlightcell 23.2 \tinystderr{1.4} & \phantom{0}7.6 \tinystderr{0.4} & \multicolumn{6}{c|}{} & 17.9 \tinystderr{0.9} \\
\ourshort[1] & 1    & 15.5  & 20.0 \tinystderr{0.6} & 15.2 \tinystderr{0.4} &  \phantom{0}6.4 \tinystderr{0.9} & \phantom{0}3.6 \tinystderr{1.3} & \multicolumn{6}{c|}{} & 11.3 \tinystderr{0.4} \\
\ourshort[2] & 2    & 17.9  & 32.4 \tinystderr{0.7} & 19.2 \tinystderr{1.2} & 10.8 \tinystderr{0.4} & \phantom{0}3.2 \tinystderr{0.4} & \multicolumn{6}{c|}{} & 16.4 \tinystderr{0.3} \\
\ourshort[4] & 4    & 22.1  & 37.2 \tinystderr{2.2} & 22.4 \tinystderr{1.5} & 14.0 \tinystderr{1.7} & \phantom{0}4.4 \tinystderr{0.7} & \multicolumn{6}{c|}{} & 19.5 \tinystderr{0.8} \\
\ourshort[8] & 8    & 31.3  & \highlightcell 44.4 \tinystderr{1.2} & 26.4 \tinystderr{0.4} & 17.2 \tinystderr{0.7} & \highlightcell \phantom{0}9.6 \tinystderr{1.0} & \multicolumn{6}{c|}{} & \highlightcell 24.4 \tinystderr{0.3} \\
\ourshort[\times] & 8 & 31.3  & 32.4 \tinystderr{0.9} & 19.6 \tinystderr{1.3} & 13.6 \tinystderr{0.9} & \phantom{0}4.8 \tinystderr{1.3} & \multicolumn{6}{c|}{} & 17.6 \tinystderr{0.5} \\
\toprule
\multicolumn{14}{c}{\textbf{RoboCasa}} \\
\midrule
Policy & 
\#Tok. & 
\makecell{Inf.\\Lat.} & 
\makecell{Close\\Drawer} & 
\makecell{Coffee Press\\Button} & 
\makecell{Turn Off\\Microwave} & 
\makecell{Turn Off\\Sink Faucet} & 
\multicolumn{6}{c|}{} & 
\makecell{Avg.} \\ 
\midrule
\diffp       & $\times$ & 35.3 & 52.0 \tinystderr{2.8} & 56.8 \tinystderr{3.6} & \highlightcell 52.8 \tinystderr{3.5} & 54.4 \tinystderr{1.8} & \multicolumn{6}{c|}{} & 54.0 \tinystderr{1.6} \\
\bin         & 384 & 888.3 & 20.4 \tinystderr{1.6} & 22.0 \tinystderr{2.3} & 31.2 \tinystderr{4.7} & 27.2 \tinystderr{3.2} & \multicolumn{6}{c|}{} & 27.7 \tinystderr{0.9} \\
\fast        & 69.7& 166.1 & 20.8 \tinystderr{2.0} & \phantom{0}8.4 \tinystderr{1.5} & 16.8 \tinystderr{1.4} & \phantom{0}6.8 \tinystderr{2.1} & \multicolumn{6}{c|}{} & 13.2 \tinystderr{1.1} \\
\quest       & 8   & 30.2 & 54.8 \tinystderr{2.1} & 55.6 \tinystderr{4.5} & 42.0 \tinystderr{2.5} & 56.8 \tinystderr{1.6} & \multicolumn{6}{c|}{} & 52.3 \tinystderr{1.9} \\
\ourshort[1] & 1   & 13.5 & 47.2 \tinystderr{3.4} & 49.6 \tinystderr{1.7} & 35.2 \tinystderr{2.6} & 58.8 \tinystderr{3.5} & \multicolumn{6}{c|}{} & 47.7 \tinystderr{1.3} \\
\ourshort[2] & 2   & 15.8 & 55.2 \tinystderr{2.2} & 53.6 \tinystderr{1.2} & 34.0 \tinystderr{4.6} & 58.4 \tinystderr{2.5} & \multicolumn{6}{c|}{} & 50.3 \tinystderr{0.8} \\
\ourshort[4] & 4   & 19.9 & 52.0 \tinystderr{1.4} & 52.8 \tinystderr{1.5} & 39.2 \tinystderr{1.2} & \highlightcell 62.8 \tinystderr{2.7} & \multicolumn{6}{c|}{} & 51.7 \tinystderr{1.0} \\
\ourshort[8] & 8   & 30.0 & 53.6 \tinystderr{2.9} & \highlightcell 63.6 \tinystderr{1.9} & 42.8 \tinystderr{3.9} & 58.4 \tinystderr{4.6} & \multicolumn{6}{c|}{} & \highlightcell 54.6 \tinystderr{1.1} \\
\ourshort[\times] & 8 & 30.0 & \highlightcell 55.6 \tinystderr{4.3} & 43.2 \tinystderr{1.9} & 36.4 \tinystderr{2.2} & 58.8 \tinystderr{1.7} & \multicolumn{6}{c|}{} & 48.5 \tinystderr{1.6} \\
\bottomrule
\end{tabular}
\caption{\textbf{Simulation benchmarking} policy performance, tokenizer compression rate (\#Tok.), and policy inference latency (Inf. Lat.) in milliseconds (ms) on one NVIDIA A100. For \fast, which generates variable-length sequences, we report the mean token count. \ourshort[K] denotes detokenization using the first $K$ tokens, while \ourshort[\times] denotes tokenizer training without nested dropout. Results report mean success rates with standard error across 5 seeds and 50 evaluation rollouts per seed per task.}
\label{appendix:table:all_sim_results}
\end{table*}

\end{document}